\documentclass[a4paper]{cas-dc}

\usepackage[numbers]{natbib}
\usepackage[shortlabels]{enumitem}
\usepackage{graphicx}
\usepackage{subcaption}
\usepackage{float}
\usepackage{placeins}
\usepackage{algorithmic}
\usepackage{algorithm}
\usepackage{array}
\hypersetup{breaklinks=true}
\def\tsc#1{\csdef{#1}{\textsc{\lowercase{#1}}\xspace}}
\tsc{WGM}
\tsc{QE}
\tsc{EP}
\tsc{PMS}
\tsc{BEC}
\tsc{DE}

\begin{document}
\let\WriteBookmarks\relax
\def\floatpagepagefraction{1}
\def\textpagefraction{.001}
\shorttitle{MARL-Based Coordinated P2P Electricity Trading}
\shortauthors{P. Wilk et~al.}

\title [mode = title]{A Hierarchical MARL-Based Approach for Coordinated Retail P2P Trading and Wholesale Market Participation of DERs}

\author[1]{Patrick Wilk}

\ead{wilkp0@rowan.edu}

\author[1]{Ethan Cantor}
\ead{cantor39@rowan.edu}

\author[2]{Yikui Liu}[orcid=0000-0002-3292-6153]
\cormark[1]   
\ead{yikuiliu@scu.edu.cn}

\author[1]{Jie Li} 
\ead{lijie@rowan.edu}

\affiliation[1]{organization={Department Electrical and Computer Engineering, Rowan University},
                addressline={201 Mullica Hill Rd}, 
                city={Glassboro},
                state={New Jersey},
                postcode={08028},
                country={United States}
                }

\affiliation[2]{organization={School of Electrical Engineering, Sichuan University},
                addressline={24 South Section 1, 1st Ring Road}, 
                city={Chengdu},
                postcode={610065}, 
                country={China}
                }

\cortext[cor1]{Corresponding author}

\begin{abstract}
The ongoing shift towards decentralization of the electric energy sector, driven by the growing electrification across end-use sectors, and widespread adoption of distributed energy resources (DERs), necessitates their active participation in the electricity markets to support grid operations. Furthermore, with bi-directional energy and communication flows becoming standard, intelligent, easy-to-deploy, resource-conservative demand-side participation is expected to play a critical role in securing power grid operational flexibility and market efficiency. This work proposes a market engagement framework that leverages a hierarchical multi-agent deep reinforcement learning (MARL) approach to enable individual prosumers to participate in peer-to-peer retail auctions and further aggregate these intelligent prosumers to facilitate effective DER participation in wholesale markets. Ultimately, a Stackelberg game is proposed to coordinate this hierarchical MARL-based DER market participation framework toward enhanced market performance.   
\end{abstract}


\maketitle

\section{Introduction}

Historically, electricity markets were structured as vertically integrated monopolies, where a single utility owned and controlled all generation, transmission, and distribution segments. As this model limited competition and consumer choice, liberalization reforms separated generation, transmission, and distribution sectors, enabling multiple large-scale suppliers to compete. Over time, however, the growing electrification and increasing integration of smaller-scale distributed energy resources (DERs) introduce new challenges to the emerging electricity market paradigm.

\subsection{Background of DERs in Wholesale and Retail Markets}

Renewable DERs such as solar photovoltaics (PV) offer environmental benefits and diversify the energy mix. While their near-zero marginal cost can lower electricity prices, their intermittency could induce higher price volatility, especially in markets with large-scale renewable adoptions \cite{BLAZQUEZ20181}. Although policies like Renewable Portfolio Standards (RPSs) have successfully accelerated decarbonization, they have also inadvertently increased retail electricity prices instead of reducing them. For instance, after 7 years of RPS implementation, retail electricity prices rose by ~11\%, partly due to indirect integration costs, stranded asset payments, and the complexity of managing intermittent supplies \cite{greenstone2020renewable, Wang2016}. Additionally, despite declining costs for both renewables and conventional generations, the U.S. retail electricity prices continued to rise, reaching 12.72 cents per kWh in 2023, a more than 85\% increase since 2000 \cite{statista_us_electricity_prices}. These increases stem from the investments in grid modernization, resiliency, and cyberphysical security measures. Given the depreciated value of U.S. electrical infrastructure at \$1.5 trillion, its replacement cost may approach \$5 trillion \cite{rhodes_us_grid_2018}. 

In parallel, the nation’s electricity sector has also seen a rapid proliferation of  consumer side DERs, such as rooftop solar, microturbines, battery storage, electric vehicles, and flexible demand resources \cite{strielkowski2021renewable}. Such DERs turn the consumers into prosumers who can generate, store, and manage their energy assets. Although prosumers can reduce their energy bills and support grid operational efficiency by contributing to local balancing, the standard top-down control and static utility tariff structures fail to harness such flexibility and complicate distribution-side coordination \cite{hahnel2020becoming}. 

Amid these challenges, local electricity markets (LEMs) emerged as a promising approach \cite{pinto_local_electricity_markets_2021, Ye_Multi_Agent_2023}. By coordinating DERs at the distribution level, participants can trade energy among themselves rather than relying on utility transactions. Such local balancing can alleviate grid congestion, delay infrastructure investment, and maximize localized renewable consumption \cite{Chen_Indirect_2019}. At scale, LEMs offer essential services to the upstream grid, providing critical support as renewables displace synchronous generators and the infrastructure becomes more stressed \cite{Strbac_Decarbonization_2021}. Within LEMs, peer-to-peer (P2P) trading is an innovative paradigm, facilitated by smart meters, communication networks, and advanced data analytics, allowing prosumers to directly negotiate and transact energy with each other \cite{Ghasemi_Mult_Agent_2020, Qiu_Scalable_2021}. With the continued development of LEMs, growing research efforts have been directed toward more effective market design and coordination frameworks:

\begin{itemize}[leftmargin=1.5em, itemsep=1pt, topsep=0pt, parsep=0pt, partopsep=0pt]

    \item \textit{From Iterative-Based to Non-Iterative Coordination of DER Aggregators:} Effective P2P markets require a robust mechanism for optimal bidding, price formation, and fair clearing to prevent market manipulation. Prosumers decide how much energy, at what price, and with which peer to trade. However, poorly coordinated P2P transactions may induce unexpected demand or generation peaks, causing grid operational challenges or system stability issues \cite{Papadaskalopoulos_Nonlinear_2016}. Aggregators have emerged to bundle DERs by pooling heterogeneous prosumers into a “virtual” resource and negotiating on their behalf in wholesale markets. Through price-based mechanisms, aggregators can harness prosumers’ flexibility to align their retail bidding strategies with wholesale market signals \cite{Yang_Deep_2022}. Trials like EcoGrid EU confirm residential loads can be price sensitive \cite{Le_Ray_Evaluating_2018}. However, research on aggregator-based coordination often relies on iterative pricing schemes, such as ADMM \cite{Khojasteh_A_Novel_2023}, that solve Nash bargaining problems by iteratively sharing price and energy data between the aggregator and participants \cite{Agwan_Pricing_2021, Wang_Incentivizing_2018}. Moreover, these models \cite{Jia_Retail_2013, Kim_An_Online_2017, Liu_Energy_Sharing_2017}, using prices to influence market behavior, can steer prosumer behavior toward the aggregator’s objectives but often require continuous, bidirectional communication, raising scalability concerns.

    \item \textit{From Centralized to Hybrid Coordination Schemes:} DER coordination in LEMs broadly follows centralized, decentralized, or hybrid approaches. Centralized models \cite{Strbac_Decarbonization_2021} typically rely on extensive information sharing by collecting all data, optimize dispatches, and manage local transactions. While this may achieve system-wide optima, it faces scalability bottlenecks, single-point failures, and privacy concerns. Decentralized models rely on bilateral contracts or consensus, alleviating scalability and privacy issues but risking suboptimal and convergence outcomes \cite{Qiu_Scalable_2021}. Hybrid methods seek to strike a balance by combining centralized coordination with decentralized decision-making, thereby retaining system-level coherence while preserving local autonomy \cite{Horrillo_Quintero_Smart_2026}.

    \item \textit{From Model-Based to Data-Driven Models:} DER coordination studies often rely on model-based optimization under ideal conditions, assuming adequate knowledge of DER characteristics and predictable prosumer behavior \cite{Liu_Energy_Sharing_2017, Horrillo_Quintero_Smart_2026, Khajeh_Local_Capacity_2021, Vinecent_Pastor_Evaluation_2019, Zhou_Framework_2020}. Although theoretically capable of delivering optimal solutions, these approaches can be computationally expensive, require extensive modeling, and become unwieldy at large scales, posing challenges for adoption by resource-limited prosumers. Data-driven methods, specifically deep reinforcement learning (DRL), have recently gained attention in energy management and market applications \cite{Guo_Change_Constrainted_2021}. Using the wealth of data, DRL agents could discover optimal decision-making policies from real or simulated experience, learning to manage uncertainties without requiring explicit system models \cite{Cao_Reinforcement_2020}. DRL has been applied to optimize residential energy trading \cite{sutton_reinforcement_learning_2018}, EV charging strategies \cite{Chen_Local_2018}, and building energy management \cite{Hua_Optimal_2019, Wan_Model_Free_2019}. However, scaling single-agent RL to a multi-agent context (MARL) is challenging due to large state-action spaces, non-stationarity, and agent heterogeneity. Current MARL falls into three categories: (i) Independent Learning \cite{Anvari_Moghaddam_A_Multi_Agent_2017, Brandi_Deep_Reinforcement_2020, Kim_Automatic_2020}, which can lead to uncoordinated outcomes and unstable training; (ii) Centralized Critic \cite{Hernandez_Leal_A_Survey_2019, Vazquez_Canteli_Multi_Agent_2019}, which mitigates non-stationarity but imposes computation and communication costs, and raises privacy issues; and (iii) Value Decomposition and Policy Sharing, suitable for cooperation but less effective when agents have divergent objectives \cite{Lu_Multi_Agent_2020}.  
\end{itemize}

Despite these developments, existing studies still largely treat retail P2P trading and wholesale market participation as two separate processes, making it difficult to capture the cross-layer interaction between local prosumer decisions and aggregator-level market actions. More specifically, a substantial proportion of studies, from the prosumer perspective, focuses on retail P2P trading mechanisms and corresponding pricing strategies, while either disregarding the utility or treating it as a fixed-price, capacity unconstrained trading counterparty for prosumers. The rest of the studies, from the aggregator perspective, focuses on pricing prosumers and participating in the wholesale market to maximize aggregator profit, while neglecting other trading channels available to prosumers. Intuitively, this separation results in inefficient coordination across market layers. Retail P2P trading decisions may deviate significantly from aggregator-level wholesale strategies, leading to economic losses for both prosumers and aggregators.

\subsection{The Proposed Solution}
The next-generation electricity market scheme, spanning both wholesale and retail layers, holds the potential to create a level playing field where all customers have equal participation opportunities. We argue that this requires intelligent, easy-to-deploy decision-making tools and involves the gamification of the energy ecosystem to stimulate adoption. This work aims to explore a market solution to streamline DER aggregator operations and enhance prosumer engagement while reducing dependence on detailed system modeling and high-bandwidth communication, the two hurdles to effective DERs’ market participation. To this end, a new market design that explicitly bridges retail P2P trading and wholesale market participation is proposed. On this basis, a hierarchical DRL framework, i.e.,the Strategic distributed Energy resource Aggregation Model for Localized Efficient System Synchronization (SEAM-LESS), combining aggregator-level Proximal Policy Optimization (PPO) policies with prosumer-level Local Strategy-Driven Multi-Agent Deep Deterministic Policy Gradient (LSD-MADDPG) coordination is proposed, enabling non-iterative coordination under limited-information sharing. 

The major contributions of this work are as follows:

\begin{enumerate}[leftmargin=*, itemsep=0pt, topsep=1pt, parsep=0pt, partopsep=0pt]
\renewcommand{\labelenumi}{(\roman{enumi})}
\setlength{\leftmargini}{-10pt}
\setlength{\labelsep}{0.5em}

  \item Effective Market Design: A new market design is proposed to unify the retail P2P trading with wholesale transactions, empowering prosumers to compete for lower costs through mutual trades while the aggregator handles unmatched supply and demand in wholesale markets. Via dynamic pricing, the aggregator fosters prosumer participation and fair bidding while competing in the wholesale market, bridging grid operation layers. This decentralized, non-iterative bidding design, guided by aggregator signals, incentivizes DERs’ P2P participation while reducing market manipulation risks through limited information sharing and centralized handling of unmatched P2P supply and demand.
  
  \item Limited-Information Sharing for Privacy-Preservation: LSD-MADDPG, a DRL solution customized for the retail P2P management, employs a decentralized actor-critic architecture that reduces non-stationarity and uncoordinated outcomes through limited information sharing, balancing cooperative and competitive objectives of individual DERs with reduced computational costs. In LSD-MADDPG, prosumer agents access only their local states and customized “strategies”, i.e., the aggregator state and price signals, during training. Such a design reduces information exchange complexity and offers a more realistic, secure, and scalable approach than fully centralized or distributed methods.
  
  \item A Hierarchical DRL Design: SEAM-LESS innovatively employs parallel PPO and LSD-MADDPG learning to coordinate DERs’ wholesale and P2P market participation. A Stackelberg game is used to model the coordination between aggregator-level PPO for wholesale market bidding with prosumer-level LSD-MADDPG for retail P2P decision-making and clearing. PPO’s stable policy updates enable the aggregator, as the Stackelberg leader, to provide consistent price signals, effectively guiding prosumers’ market behaviors.  
  
\end{enumerate}

The remainder of this paper is organized as follows: Section 2 introduces the hierarchical DER aggregation framework, detailing the market rules and pricing structures. Section 3 formulates the Markov game for aggregator and prosumer coordination and the proposed Stackelberg game, defining SEAM-LESS architecture. Section 4 conducts case studies, discussing simulation results and comparative analyses of the proposed with a conventional centralized critic MADDPG and rule-based market solutions. Finally, Section 5 concludes with key insights and future research directions. 

\vspace{-5pt}
\section{Hierarchically Coordinated Wholesale and P2P Market Participation of DERS}

\begin{figure}[pos=t]
\centering
\begin{subfigure}[b]{0.3\textwidth}
    \centering
    \includegraphics[width=\textwidth]{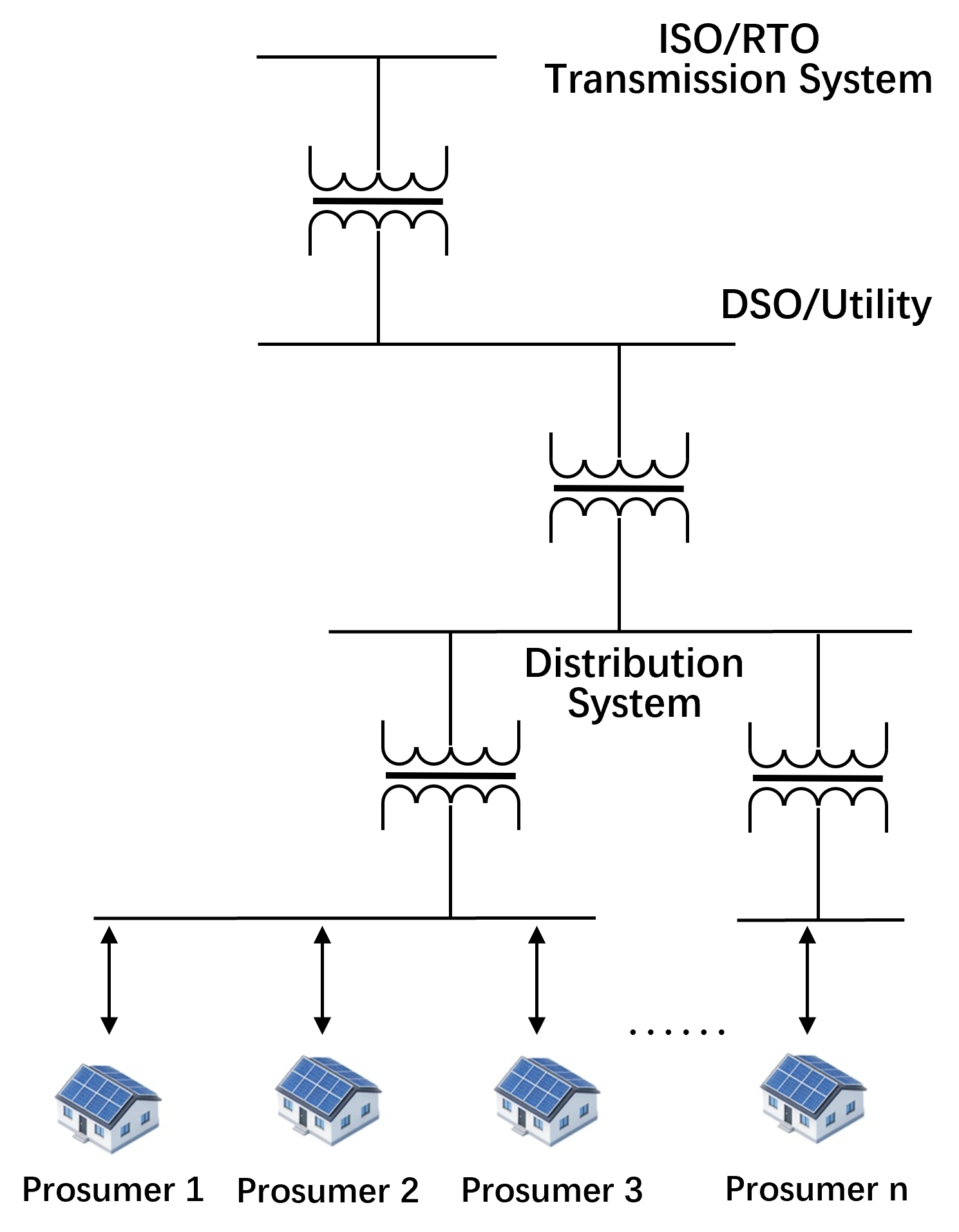}
    \caption{Physical Network Structure}
    \label{fig_1a}
\end{subfigure}

\par\vspace{6pt}
\begin{subfigure}[b]{0.4\textwidth}
    \centering
    \includegraphics[width=\textwidth]{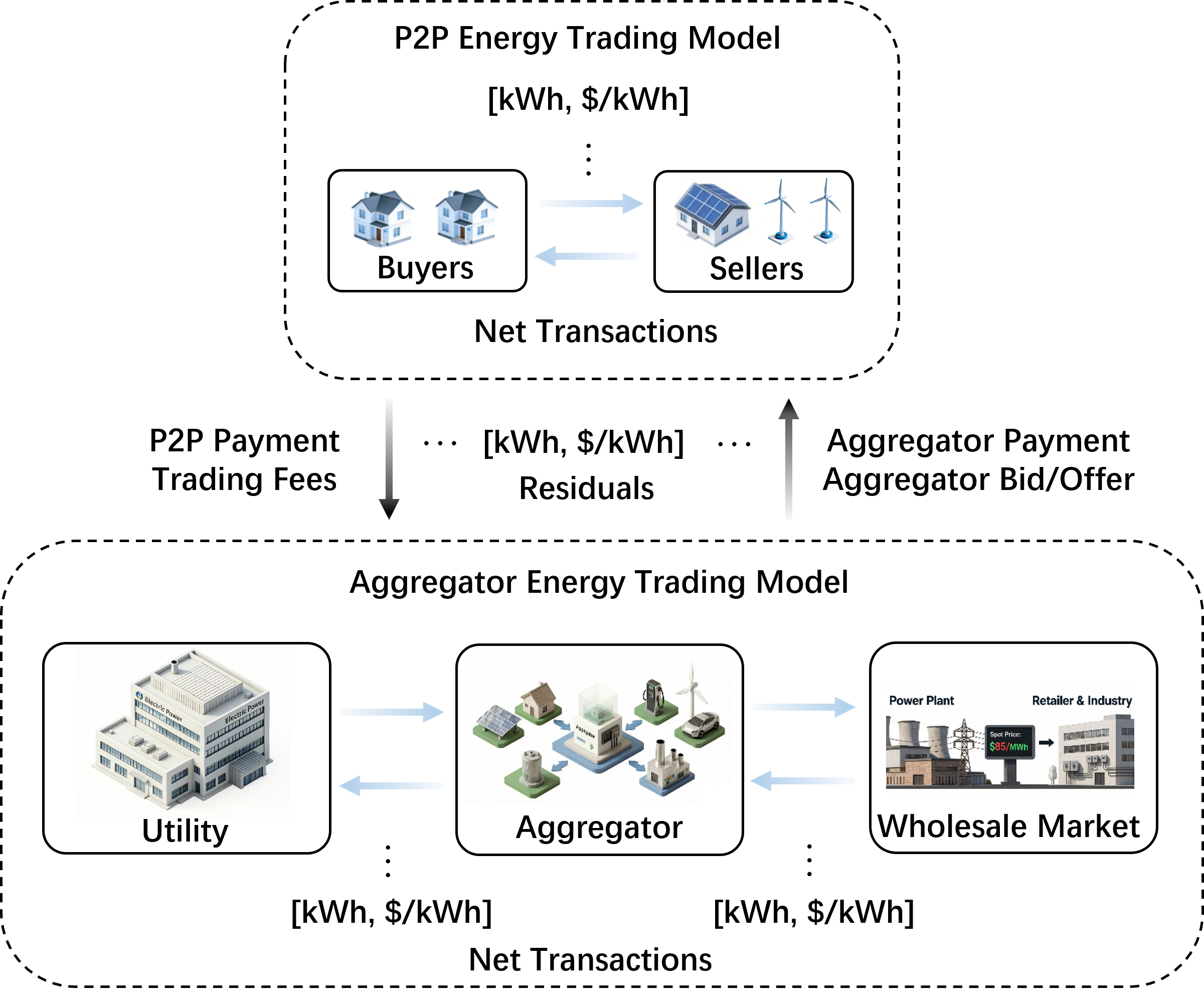}
    \caption{Hierarchical Market Structure}
    \label{fig_1b}
\end{subfigure}

\caption{Proposed Wholesale and P2P Market Participation Framework of DERs.}
\vspace{-20pt}
\label{fig_1}
\end{figure}

Figure \ref{fig_1} illustrates the physical and market structures of the proposed hierarchical DER market participation framework. In Figure. \ref{fig_1a}, the physical network shows how electric energy flows between DERs/prosumers connected to the distribution and transmission grid. In Figure. \ref{fig_1b}, the hierarchical wholesale-retail market reveals how the prosumers, aggregator, utility, and ISO/RTO interact via digital platforms. Specifically, the DER aggregator interfaces with the retail P2P market, utility, and wholesale market, facilitating transactions in terms of energy quantity and price. Decoupling the physical and financial layers promotes market functioning via digital clearinghouses while maintaining the grid in its technical limits.   

In the retail market, the P2P protocol matches bids and offers to form financial agreements among prosumers, while the aggregator observes prosumer consumption and generation patterns, adapts its wholesale market strategies, and exploits price differentials across market layers. Through these actions, the aggregator promotes local energy exchanges, coordinates dynamically with the wholesale market, and seeks to profit. 

The P2P market aims to maximize the combined net benefits of buyers in set $X$ and sellers in set $Y$ (\ref{Eq_1}). Each buyer $x$ derives value $V_x(Q_x)$ from the energy it purchases, while each seller $y$ incurs cost $C_y(Q_y)$ for producing the energy it sells. Each transaction $Q_{x,y}$ between buyer $x$ and seller $y$ has a price $P_{x,y}$.

\vspace{-10pt}
\begin{equation}
\begin{aligned}
 \max 
\sum\nolimits_{x \in X} \left[ V_x(Q_x) - \sum\nolimits_{y \in Y} P_{x,y} \cdot Q_{x,y} \right] \\
 + \sum\nolimits_{y \in Y} \left[ \sum\nolimits_{x \in X} P_{x,y} \cdot Q_{x,y} - C_y(Q_y) \right]
 \label{Eq_1}
\end{aligned}
\end{equation}

\noindent where $Q_x=\sum_{y\in Y}Q_{x,y} $  is the total quantity purchased by buyer $x$ and $Q_y=\sum_{x\in X}Q_{x,y}$ is the total quantity sold by seller $y$.

The DER aggregator seeks to maximize its own profit (\ref{Eq_2}) by managing unsettled bids/offers from the P2P market via wholesale transactions at price $p_a^w$. The aggregator sells energy to the P2P market at price $p_a^s$  and acquires surplus at $p_a^b$. Unsettled supply from P2P sellers is $Q_{a,y}$, and unsettled demand from P2P buyers is $Q_{a,x}$.

\vspace{-5pt}
\begin{equation}
\begin{aligned}
\max \sum\nolimits_{x \in X} p_a^s \cdot Q_{a,x}
- \sum\nolimits_{y \in Y} p_a^b \cdot Q_{a,y}\\
- p_a^w \left( \sum\nolimits_{x \in X} Q_{a,x} - \sum\nolimits_{y \in Y} Q_{a,y} \right) 
\label{Eq_2}
\end{aligned}
\end{equation}
\vspace{-15pt}

\subsection{P2P Retail Market Design}
The retail P2P market is executed in discrete, equal time intervals. During each time interval, participants submit their bids/offers, and the P2P market settles all trades in a single round before proceeding to the next time interval. A P2P market participant could declare itself as a buyer or a seller based on its assets’ evolving conditions (e.g., generation/storage capacity, demand flexibility) and strategic considerations at each time interval. The P2P market design is elaborated via the following three steps, where $n$ indexes prosumers (buyers and sellers) in the P2P market participant set $N=N_{B}\cup N_{O}$, with $N_B$ being the buyer set, $N_O$ being the seller set, and $N_{B}\cap N_{O}=\varnothing$.  

\textit{Step 1: Bid/Offer Submission and Sorting.} Each buyer $n\hspace{-1mm}\in \hspace{-1mm}N_B$ submits a bid $b_n$ characterized by a maximum price $p_{b_{n}}$, at which it is willing to pay for a desired quantity $q_{b_{n}}\hspace{-1mm}<0$ (negative denotes energy purchasing). Similarly, each seller $n\hspace{-0.5mm}\in \hspace{-0.5mm}N_O$ submits an offer $o_n$ with a minimum acceptable price $p_{o_{n}}$ for quantity $q_{o_{n}}\hspace{-0.5mm}>0$ (positive denotes energy selling). All bids are collected in (\ref{Eq_3}) and all offers in (\ref{Eq_4}).

\vspace{-10pt}
\begin{equation}
\qquad B = \left\{ b_n \left( p_{b_n}, q_{b_n} \right) \,\middle|\, n \in N_B \right\}
\label{Eq_3}
\end{equation}

\vspace{-5pt}
\begin{equation}
\qquad O = \left\{ o_n \left( p_{o_n}, q_{o_n} \right) \,\middle|\, n \in N_O \right\}
\label{Eq_4}
\end{equation}

Once all bids and offers are submitted, the P2P platform sorts all bids in descending price order (\ref{Eq_5}), and all offers in ascending price order (\ref{Eq_6}), enabling efficient matching of the highest-paying buyers with the lowest-priced sellers.

\vspace{-15pt}
\begin{equation}
\hspace{-3mm }B_k^* = \left\{ b_{1,k}^*, b_{2,k}^*, \ldots, b_{n_b,k}^* \right\}
\ \hspace{-1mm }\text{with} \
p_{b_{1,k}^*} \hspace{-1mm } \ge p_{b_{2,k}^*} \hspace{-1mm }\ge \hspace{-1mm }\cdots \hspace{-1mm }\ge p_{b_{n_b,k}^*}
\label{Eq_5}
\end{equation}

\vspace{-20pt}
\begin{equation}
\hspace{-4mm }O_k^* = \left\{ o_{1,k}^*, o_{2,k}^*, \ldots, o_{n_o,k}^* \right\}
\ \hspace{-1mm }\text{with} \
p_{o_{1,k}^*} \hspace{-1mm }\le p_{o_{2,k}^*} \hspace{-1mm }\le \hspace{-1mm }\cdots \hspace{-1mm }\le \hspace{-1mm } p_{o_{n_b,k}^*}
\label{Eq_6}
\end{equation}
\vspace{-10pt}

Here, $n_b =|N_B|$ and $n_o =|N_O|$ represent the number of bids and offers in current time interval. The superscript (*) denotes the sorted lists, and $k$ indexes the round of the matching process. $b_{i,k}^*$ is the $i$-th highest-priced bid and $o_{j,k}^*$ is the $j$-th lowest priced offer at round $k$.

\textit{Step 2: Price Matching and Proportional Allocation.} For the $k$-th round, identify:
\begin{itemize}[itemsep=0pt, topsep=1pt, parsep=0pt, partopsep=0pt]
    \item 	The highest bid $b_k^*= b_{1,k}^*$ with $p_{b_k^*}=p_{b_{1,k}^*}$, $q_{b_k^*}=q_{b_{1,k}^*}$
    \item The lowest bid $o_k^*= o_{1,k}^*$ with $p_{o_k^*}=p_{o_{1,k}^*}$, $q_{o_k^*}=q_{o_{1,k}^*}$
\end{itemize}

If $p_{b_k^*}\geq p_{o_k^*}$, a buy/sell match is possible, meaning at least one buyer is willing to pay no lower than the seller requires; otherwise, no transaction occurs. 

For multiple buyers or sellers sharing the same price, we group them as (\ref{Eq_7}) and (\ref{Eq_8}). All buyers in $B_{p,k}$ have the same highest price at round $k$, the total demand at this price is calculated as (\ref{Eq_9}), and each buyer’s proportion of the total demand at this price is (\ref{Eq_10}). Similarly, all sellers in $O_{p,k}$ share the same lowest price at round $k$, the total supply is the summation (\ref{Eq_11}), and each seller’s proportion of the total supply at this price is (\ref{Eq_12}). 

\vspace{-10pt}
\begin{equation}
 B_{p,k} = \left\{ b_{i,k}^* \in B_k^* \,\middle|\, p_{b_{i,k}^*} = p_{b_k^*} \right\}
\label{Eq_7}
\end{equation}

\begin{equation}
 O_{p,k} = \left\{ o_{j,k}^* \in O_k^* \,\middle|\, p_{o_{j,k}^*} = p_{o_k^*} \right\}
\label{Eq_8}
\end{equation}

\begin{equation}
 Q_{TD}^{(k)} = \sum\nolimits_{b_{i,k}^* \in B_{p,k}} \left( - q_{b_{i,k}^*} \right)
\label{Eq_9}
\end{equation}

\begin{equation}
 \alpha_{b_{i,k}^*}^{(k)} = - q_{b_{i,k}^*}/Q_{TD}^{(k)}
\label{Eq_10}
\end{equation}

\begin{equation}
 Q_{TS}^{(k)} = \sum\nolimits_{o_{j,k}^* \in O_{p,k}} q_{o_{j,k}^*}
\label{Eq_11}
\end{equation}

\begin{equation}
 \beta_{o_{j,k}^*}^{(k)} = q_{o_{j,k}^*}/Q_{TS}^{(k)}
\label{Eq_12}
\end{equation}

The total transaction quantity that can be matched at round $k$ is limited by the smaller of total demand and total supply (\ref{Eq_13}). $q_k$  is then allocated proportionally to individual participants for updated quantities for buyers (\ref{Eq_14}) and sellers (\ref{Eq_15}).

\vspace{-10pt}
\begin{equation}
 q_k = \min \left( Q_{TD}^{(k)}, \, Q_{TS}^{(k)} \right)
\label{Eq_13}
\end{equation}

\vspace{-5pt}
\begin{equation}
q_{b_{i,k}^*}^{tx,(k)} = \alpha_{b_{i,k}^*}^{(k)} \times q_k,
\quad \forall \, b_{i,k}^* \in B_{p,k}
\label{Eq_14}
\end{equation}

\vspace{-5pt}
\begin{equation}
q_{o_{j,k}^*}^{tx,(k)} = \beta_{o_{j,k}^*}^{(k)} \times q_k,
\quad \forall \, o_{j,k}^* \in O_{p,k}
\label{Eq_15}
\end{equation}

The transaction price $p_k$ is set as the midpoint between the highest buyer price and the lowest seller price (\ref{Eq_16}). This design ensures fairness, as buyers pay less than or equal to their maximum willingness-to-pay, and sellers receive at least their minimum acceptable price.

\vspace{-5pt}
\begin{equation}
p_k = (p_{b_k^*} + p_{o_k^*})/2
\label{Eq_16}
\end{equation}

Finally, buyer payments and seller revenues for round $k$ are (\ref{Eq_17}) and (\ref{Eq_18}), respectively.

\vspace{-5pt}
\begin{equation}
PA(b_{i,k}^*)^{(k)} = q_{b_{i,k}^*}^{tx,(k)} \times p_k
\label{Eq_17}
\end{equation}

\vspace{-5pt}
\begin{equation}
RV(o_{j,k}^*)^{(k)} = q_{o_{j,k}^*}^{tx,(k)} \times p_k
\label{Eq_18}
\end{equation}

The remaining quantities for each participant are updated with transacted amounts (\ref{Eq_19}) and (\ref{Eq_20}), and any fully cleared bids/offers (e.g., $q_{b_{i,k}^*} \hspace{-1mm}=0$ or $q_{o_{j,k}^* }\hspace{-1mm}=0$) are removed from the active lists. $PA_n$ and $RV_n$ are updated via (\ref{Eq_21}) and (\ref{Eq_22}) to record the total amount paid by a buyer and the total revenue earned by a seller across all P2P transactions up to round $k$.

\vspace{-10pt}
\begin{equation}
q_{b_{i,k}^*} \leftarrow q_{b_{i,k}^*} + q_{b_{i,k}^*}^{tx,(k)}
\label{Eq_19}
\end{equation}

\vspace{-5pt}
\begin{equation}
q_{o_{j,k}^*} \leftarrow q_{o_{j,k}^*} + q_{o_{j,k}^*}^{tx,(k)}
\label{Eq_20}
\end{equation}

\vspace{-5pt}
\begin{equation}
PA_n \leftarrow PA_n + PA(b_{i,k}^*)^{(k)},
\quad n\in N_B
\label{Eq_21}
\end{equation}

\vspace{-10pt}
\begin{equation}
RV_n \leftarrow RV_n + RV(o_{j,k}^*)^{(k)},
\quad n\in N_O
\label{Eq_22}
\end{equation}

After adjusting the lists, set $k=k+1$ and repeat this matching process until $p_{b_k^*} <p_{o_k^*}$, or one list is empty.

\textit{Step 3: Unmatched Quantities.} After all P2P transactions are cleared, some bids and/or offers may not be fully matched, and will proceed to the aggregator for further settlement. 

\subsection{Aggregator’s Wholesale Market Participation}

During a time interval, each participant’s unmatched quantity $q_{b_n}^a$ or $q_{o_n}^a$  is handled by the aggregator, which sums them as in (\ref{Eq_23}) for wholesale trading. 

\vspace{-5pt}
\begin{equation}
q_a^w = \sum\nolimits_{n \in N_B} q_{b_n}^a + \sum\nolimits_{n \in N_O} q_{o_n}^a
\label{Eq_23}
\end{equation}

If $q_a^w>0$, the aggregator is a net seller during that time interval in the wholesale market and otherwise, a net buyer. The aggregator chooses a price $p_a^w$ at which it attempts to transact in the wholesale market. To be a successful wholesale seller, the aggregator must offer a price no higher than the wholesale clearing price; and for a successful purchase, it must bid no lower than the wholesale clearing price. If the aggregator fails to clear in the wholesale market, it incurs a penalty paid to the P2P participants, proportional to their unfulfilled quantities $q_{o_n}^a$ and $q_{b_n}^a$. The aggregator must then buy from or sell to the utility at preset prices on behalf of these participants. If successful, the aggregator transacts at prices relative to the wholesale clearing price $P_w$ to buy (\ref{Eq_24}) and sell (\ref{Eq_25}) unmatched P2P quantities. $\Delta_b$ and $\Delta_o$ represent revenue margins around the wholesale price, granting the aggregator flexibility to profit by acquiring P2P energy below $P_w$ and selling above it.

\vspace{-5pt}
\begin{equation}
p_a^b = P_w \times \left( 1 - \Delta_b \right)
\label{Eq_24}
\end{equation}

\vspace{-10pt}
\begin{equation}
p_a^s = P_w \times \left( 1 + \Delta_o \right)
\label{Eq_25}
\end{equation}

The payments of the P2P buyer to and revenue of the P2P seller from the aggregator are (\ref{Eq_26}) and (\ref{Eq_27}), respectively. 

\vspace{-5pt}
\begin{equation}
PA_n^a = q_{b_n}^a \times p_a^s,
\quad n\in N_B
\label{Eq_26}
\end{equation}

\vspace{-7pt}
\begin{equation}
RV_n^a = q_{o_n}^a \times p_a^b,
\quad n\in N_O
\label{Eq_27}
\end{equation}

If the aggregator fails to clear in the wholesale market, it pays penalty $PN_n$, proportional to the unfulfilled quantities (\ref{Eq_28}-\ref{Eq_29}).

\vspace{-15pt}
\begin{equation}
PN_n = p_p^a \, \cdot \big( - q_{b_n}^a \big), 
\quad n \in N_B
\label{Eq_28}
\end{equation}

\vspace{-7pt}
\begin{equation}
PN_n = p_p^a \, \cdot \big( q_{o_n}^a \big), 
\quad n \in N_O
\label{Eq_29}
\end{equation}

The aggregator’s profit (\ref{Eq_30}) includes payments from/to P2P buyers/sellers, transactions settled in the wholesale market, and penalties incurred if the wholesale market attempt fails. 

\vspace{-15pt}
\begin{equation}
TA = 
- \sum\nolimits_{n \in N_B} \hspace{-1mm}PA_n^a
- \sum\nolimits_{n \in N_O} \hspace{-1mm}RV_n^a
+ \left( P_w \hspace{-0.5mm}\cdot \hspace{-0.5mm}q_a^w \right)
- \sum\nolimits_n PN_n
\label{Eq_30}
\end{equation}

Final payouts of P2P buyer (\ref{Eq_31}) and seller (\ref{Eq_32}) incorporate P2P transactions and settlements with the aggregator.

\vspace{-5pt}
\begin{equation}
TB_n = PA_n + PA_n^a + PN_n,
\quad n \in N_B
\label{Eq_31}
\end{equation}

\vspace{-10pt}
\begin{equation}
TS_n = RV_n + RV_n^a + PN_n,
\quad n \in N_O
\label{Eq_32}
\end{equation}

The proposed hierarchical market framework spans a 24-hour horizon. For each hour $t$, the aggregator posts its purchase and sell prices,  $p_a^b(t)$ and $p_a^s(t)$, to the P2P market before participants submit bids and offers for that hour. After the P2P market clears, settled prices and quantities are allocated among participants, with any unmatched bids and/or offers tracked as residuals. Once all 24 hours’ P2P transactions are processed, the aggregator compiles hourly residuals and transacts with the wholesale market or utility on behalf of prosumers, aiming to optimize its profit (\ref{Eq_30}). Following wholesale settlements, the aggregator distributes net profits or costs back to the P2P participants. This approach accommodates the single round 24-hour day-ahead wholesale market while allowing hourly price-setting between the aggregator and local P2P participants.

\vspace{-5pt}
\section{DRL-Based Stackelberg Game Framework For Market Interactions}
The above proposed retail P2P market and the aggregator’s participation in the wholesale market can be formulated as a Markov game. In this setting, both individual prosumers (i.e., P2P market participants) and the aggregator operate as strategic RL agents, making decisions along with evolving system and market states. This captures dynamic behaviors and interactions between the aggregator and participants and among participants. We first describe the Markov game formulations for aggregator and prosumer coordination, then present the proposed SEAM-LESS architecture based on the Stackelberg game framework.

\subsection{Markov Game for Market Interactions}
\subsubsection{P2P Market Agents State and Action Space}
The state space of a P2P market agent includes all variables influencing operational and market dynamics:

\vspace{-5pt}
\begin{equation}
s_n = \{ p_a^b, \, p_a^s, \, q_n^{\mathrm{tot}}, \, p_n^{\mathrm{avg}}, \, q_{\mathrm{cm}}, \, TP_n \}
\label{Eq_33}
\end{equation}

For agent $n$, the total quantity $q_{n}^{tot}$ is $q_{b_n} $ if it is a buyer or $q_{o_n}$ if it is a seller. The total payoff (or profit) $TP_n$  is defined in (\ref{Eq_34}). The average P2P trade price of agent $n$ is $p_{n}^{avg}$ (\ref{Eq_35}), and the aggregated total net quantity of all prosumers is $q_{cm}$ (\ref{Eq_36}).

\vspace{-10pt}
\begin{equation}
TP_n =
\begin{cases}
TB_n, & \text{if } n \in N_B \\
TS_n, & \text{if } n \in N_O
\end{cases}
\label{Eq_34}
\end{equation}

\vspace{-5pt}
\begin{equation}
p_n^{avg} = TP_n/q_n^{tot}
\label{Eq_35}
\end{equation}

\vspace{-10pt}
\begin{equation}
q_{cm} = \sum\nolimits_n q_n^{tot}
\label{Eq_36}
\end{equation}

Agent $n$’s action space $a_n  = [a_n^p]$ is designed to adjust its bid/offer within predefined bounds $[p_{min}, p_{max}]$ (\ref{Eq_37}) 
where $a_n^p \in [-1,1]$ represents a normalized action.

\vspace{-15pt}
\begin{equation}
p_{b_n}/p_{o_n} \hspace{-0.75mm}=\hspace{-0.5mm} ((a_n^p \hspace{-0.5mm}+\hspace{-0.5mm} 1)/2) \cdot(p_{\max} \hspace{-0.5mm}-\hspace{-0.5mm} p_{\min}) + p_{\min},
\, n \hspace{-0.5mm}\in\hspace{-0.5mm} N_B/N_O
\label{Eq_37}
\end{equation}

\subsubsection{Aggregator Agent State and Action Space}
The aggregator agent acts as an intermediary between the P2P and wholesale markets, aiming to align local trading with broader system objectives and wholesale price signals. Its state space is defined as (\ref{Eq_39}).

\vspace{-10pt}
\begin{equation}
s_{ag} = \{ F_{MP}, \, F_{IP}, \, q_{a,o}^w, \, q_{a,b}^w \}
\label{Eq_39}
\end{equation}

$F_{MP}$ provides the aggregator with insight into the expected wholesale clearing price in the current trading period, while the subsequent period’s wholesale clearing price forecast $F_{IP}$ further guides the aggregator’s buy/sell price settings to shape its behavior in the retail P2P market. The aggregated supply $q_{a,o}^w$ (\ref{Eq_40}) and demand $q_{a,b}^w$ (\ref{Eq_41}) from the P2P market define the net local energy balance to be managed by the aggregator:  

\vspace{-5pt}
\begin{equation}
q_{a,o}^w = \sum\nolimits_{n \in N_O} q_{o_n}^a
\label{Eq_40}
\end{equation}

\vspace{-5pt}
\begin{equation}
q_{a,b}^w = \sum\nolimits_{n \in N_B} q_{b_n}^a
\label{Eq_41}
\end{equation}

The aggregator’s action space (\ref{Eq_42}) includes setting the wholesale market bid and adjusting its P2P buy/sell prices:

\vspace{-5pt}
\begin{equation}
a_{ag} = \{ a_{ag}^{w}, \, a_{ag}^{b}, \, a_{ag}^{o} \}
\label{Eq_42}
\end{equation}

The wholesale bidding price $p_a^w$ (43) is constrained between $p_{ag}^{min}$ and $p_{ag}^{max}$ via the normalized action $a_{ag}^{w} \in [0,1]$.

\vspace{-5pt}
\begin{equation}
p_{a}^{w} =  p_{ag}^{max} \cdot a_{ag}^{w} + p_{ag}^{min}
\label{Eq_43}
\end{equation}

The aggregator’s buy action $a_{ag}^b \in [0,1]$ and offer action $a_{ag}^o \in [0,1]$ in P2P market are defined as fractions of the maximum allowable markup $\rho^{max}$  applied to the forecasted wholesale price $F_{MP}$. The adjustments $\Delta_b$ and $\Delta_o$, used to determine the aggregator’s buy and sell prices in the P2P market (as defined in (\ref{Eq_24}) and (\ref{Eq_25})), are then given by (\ref{Eq_44}) and (\ref{Eq_45}). 

\vspace{-15pt}
\begin{equation}
\Delta_b = a_{ag}^b \cdot \rho^{\max}
\label{Eq_44}
\end{equation}

\vspace{-10pt}
\begin{equation}
\Delta_o = a_{ag}^o \cdot \rho^{\max}
\label{Eq_45}
\end{equation}

\vspace{-10pt}
\subsubsection{Reward Structures}
The base reward (\ref{Eq_46}) quantifies each agent’s cost savings or revenue via P2P trading compared to the aggregator fallback. A P2P participant’s cost or revenue ($PA_n$  or $RV_n$) is compared to that it would have obtained from the aggregator for the portion traded in the P2P market.

\vspace{-15pt}
\begin{equation}
r_n^* =
\begin{cases}
RV_n - (q_{o_n} - q_n^{un}) \times p_a^b, & \text{if } n \in N_O \\[2mm]
- PA_n - (-q_{b_n} + q_n^{un}) \times p_a^s, & \text{if } n \in N_B
\end{cases}
\label{Eq_46}
\end{equation}
\vspace{-10pt}

To ensure continuous feedback from action selection, a penalty $\Pi_n$ is introduced when the P2P participant’s unmatched energy $q_{n}^{un}$ stems from an uncompetitive P2P bid/offer (\ref{Eq_47}-\ref{Eq_49}). This helps agents detect uncompetitive pricing, balancing exploration (agents can safely try moderate price variations) with exploitation (excessively high or low prices incur aggregator fallback). The result is a clear signal reflecting the “missed opportunity” of potential savings. The final agent reward $r_n$ is the base reward minus penalty (\ref{Eq_50}). If no trade occurs, $r_n^*= 0$.

\vspace{-15pt}
\begin{equation}
\Delta_n^S = \max \left( 0, \, p_{o_n} - p_a^b \right),
\Pi_n^S = q_{o_n}^a \times \Delta_n^S, 
\quad n \in N_O
\label{Eq_47}
\end{equation}

\vspace{-20pt}
\begin{equation}
\Delta_n^B = \max \left( 0, \, p_a^s - p_{b_n}   \right),
\Pi_n^B = q_{b_n}^a \times \Delta_n^B, 
\quad n \in N_B
\label{Eq_48}
\end{equation}

\vspace{-10pt}
\begin{equation}
\Pi_n =
\begin{cases}
\Pi_n^S, & \text{if } n \in N_O \\
\Pi_n^B, & \text{if } n \in N_B
\end{cases}
\label{Eq_49}
\end{equation}

\vspace{-5pt}
\begin{equation}
    r_{n} = r_{n}^{*} - \Pi_{n}
\label{Eq_50}
\end{equation}

The aggregator reward $r_{ag}$ depends on successful arbitrage and alignment with wholesale market conditions (\ref{Eq_51}), representing net gains from wholesale transactions and P2P redistributions.

\vspace{-15pt}
\begin{equation}
    r_{ag} = TA
\label{Eq_51}
\end{equation}

\subsection{Hierarchical DRL-Based Stackelberg Game Framework}
\begin{figure}[pos=h]
\centering
\vspace{-5pt}
\includegraphics[width=0.4\textwidth]{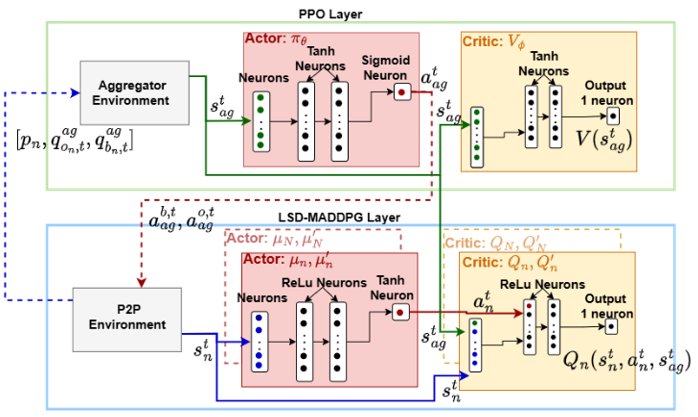}
\vspace{-5pt}
\caption{SEAM-LESS Stackelberg Game Framework.}
\vspace{-10pt}
\label{fig_2}
\end{figure}

A hierarchical DRL framework, i.e., SEAM-LESS,  is proposed under a Stackelberg game setup to solve the Markov game, as shown in Figure \ref{fig_2}. The aggregator agent, as the game leader, sets price signals to define its wholesale and retail market policies, while the prosumer agents, as the game followers, learn local control policies
to govern their decision-making in the retail market. 

To establish a fair and privacy-preserving retail market, prosumers are designed with limited access to the information of their peers. Although a centralized operator with full observability could solve the retail market clearing problem, it inevitably compromises the privacy of market participants, lacks scalability, and faces increased computational complexity as the number of P2P participants increases. Instead, LSD-MADDPG, a model-free MARL scheme proposed in \cite{Lu_Multi_Agent_2020}, is adopted and customized.

The proposed SEAM-LESS framework employs PPO for the aggregator and LSD-MADDPG for prosumers, leveraging their suitability for a non-iterative, hierarchical DRL approach in a Stackelberg game structure. In practical wholesale markets, bids are submitted once per period (e.g., hourly) and remain fixed, aligning with RL’s single-action policies that ensure timely, computationally efficient decisions, unlike iterative methods that risk manipulation through intra-period adjustments. This design is adopted in the proposed hierarchical wholesale-retail framework, where PPO’s clipped surrogate objective provides stable price signals, critical for guiding prosumer behavior as the Stackelberg leader, while its proficiency in continuous action spaces and high-dimensional inputs supports precise wholesale bidding. LSD-MADDPG’s decentralized architecture limits information sharing to local observations and aggregator signals, unlike conventional MADDPG, which assumes that each agent’s critic accesses all other agents’ observations and actions, ensuring privacy and scalability. Furthermore, its ability to balance cooperative and competitive behaviors enables effective P2P coordination, validated by the authors’ prior success in community energy management \cite{Wilk_Multi_Agent_2024}. In LSD-MADDPG’s actor-critic scheme, prosumer agents’ critics use their local observations and shared “strategies” from the aggregator to facilitate coordinated policy learning. This design respects practical physical constraints and upholds data confidentiality, critical to an effective retail energy market. By designing rewards to align individual prosumer goals with the aggregator’s high-level incentives, LSD-MADDPG ensures that local decisions collectively yield successful P2P settlements.  

During training, the aggregator agent broadcasts its price signals globally to the P2P market as a game leader using a PPO policy, while LSD-MADDPG agents, as game followers, observe their own local states and aggregator’s signals and state to make decisions. This design ensures that the aggregator maintains full transparency, a fundamental principle for an open and effective P2P market, while helping prosumers understand market conditions and the impact of their actions on market clearing. Consequently, it reduces market manipulation and promotes rational bidding, enhancing P2P market stability and integrity. By combining cooperative (agents collectively responding to aggregator signals and state) and competitive (each agent pursues individual market gains) market behaviors, SEAM-LESS excels in this mixed MARL environment. 

As shown in Algorithm \ref{alg:SEAMLESS_Workflow}, the aggregator’s PPO policy operates at a higher tier as the leader. The aggregator observes the wholesale and retail market conditions and sets a wholesale market bid/offer price $a_{ag}^{w}$ and P2P retail market incentive signals $[a_{ag}^b,a_{ag}^o]$. These signals are broadcasted to all the prosumers in the P2P market, which run LSD-MADDPG follower policies. Each LSD-MADDPG agent receives its local observation $s_n$ and the aggregator’s updated state $s_{ag}$, enabling it to determine actions $a_n$ that adjust its P2P trading profile. Any residual energy not settled in the P2P market is managed by the aggregator. A dynamic feedback loop emerges as the wholesale transaction success of the aggregator is dependent on the responsiveness of all the LSD-MADDPG agents. This closed-loop interaction fosters a self-regulating market where agents learn to trade optimally under evolving incentives.

\begin{algorithm}[!h]
\footnotesize
\caption{Training of SEAM-LESS}
\label{alg:SEAMLESS_Workflow}
\begin{algorithmic}[1]

\STATE \textbf{Initialize each LSD-MADDPG agent $n$ and PPO agent:} \\
\begin{itemize}[leftmargin=1em, itemsep=0pt, topsep=0pt, parsep=0pt, partopsep=0pt]
    \item Initialize actor network $\mu_{n}$ with $\theta_{n}^{\mu_n}$ and critic network $Q_n$ with $\theta_{n}^{Q_n}$;
    \item Initialize target networks $\mu_n'$ and $Q_n'$ with weights $\theta_{n}^{\mu_n'}\leftarrow \theta_{n}^{\mu_n}$ and $\theta_{n}^{Q_n'}\leftarrow \theta_{n}^{Q_n}$;
    \item Initialize LSD experience replay buffer $\mathcal{D}$;
    \item Initialize PPO parameters, including actor network $\pi_{\theta}$ and critic network $V_{\phi}$. PPO uses its own data buffer for trajectories.
\end{itemize}

\STATE \textbf{Training Loop for each episode $e$ = 1,2,…:}\\
\begin{itemize}[leftmargin=1em, itemsep=0pt, topsep=0pt, parsep=0pt, partopsep=0pt]
    \item Reset the environment: get initial observation $s_n^0$ for each LSD-MADDPG agent $n$, and get $s_{ag}^0$ initial state for PPO.
    \item  \textbf{For} each timestep $t$ until the terminal state:
    \begin{enumerate}[label=\Alph*., leftmargin=1.5em, itemsep=0pt, topsep=0pt, parsep=0pt, partopsep=0pt]
        \item LSD-MADDPG Agents’ Actions: Each agent $n$  observes $(s^t_n, s^t_{ag})$. Select action $a^t_n=\mu_n(o^t_n)$.
        \item Partial Environment Step:
        \begin{enumerate}[label=i., leftmargin=1em, itemsep=0pt, topsep=0pt, parsep=0pt, partopsep=0pt]
            \item Execute actions $\left\{ a^t_n \right\}$ in LSD-MADDPG layer;
            \item Update $s_{ag}^t$ resulting in $s_{ag}^{t,*}$.
        \end{enumerate}
        \item PPO Actions: PPO observes $s_{ag}^{t,*}$. Select action $a_{ag}^t\sim \pi_\theta (a_{ag}^t |s_{ag}^{t,*} )$.
        \item Finalized Environment Step:
        \begin{enumerate}[label=i., leftmargin=1em, itemsep=0pt, topsep=0pt, parsep=0pt, partopsep=0pt]
            \item Execute actions $a_{ag}^t$. Next states for LSD $s_{n}^{t'}$ and PPO $s_{ag}^{t'}$;
            \item Rewards  $r_{n}^t$ and $r_{ag}^t$. Get terminal signals \textit{done}.
        \end{enumerate}
        \item Updates for LSD-MADDPG and PPO:
        \begin{enumerate}[label=i., leftmargin=1em, itemsep=0pt, topsep=0pt, parsep=0pt, partopsep=0pt]
            \item Every $\delta$ steps for LSD-MADDPG sample:
            \begin{itemize}[leftmargin=1em, itemsep=0pt, topsep=0pt, parsep=0pt, partopsep=0pt]
                \item 	Batch of $\rho$ transitions $(s_n^j, a_n^j,r_n^j,s_n^{j' },s_{ag}^j,s_{ag}^{j'})$ from $\mathcal{D}$;
                \item 	Compute critic loss: $L(\theta_n^{Q_n} )=  \frac{1}{\rho} \sum_{j=1}^{\rho}(y_n^j  -Q_n (s_n^j,a_n^j,s_{ag}^j))^2$;
                \item Perform gradient descent on $L(\theta_n^{Q_n} )$ to update $\theta_{n}^{Q_n}$;
                \item Compute policy gradient: $\nabla_{\theta^{\mu_n}}  J({\theta_n^{\mu_n}}) = \mathbb{E}[\nabla_{\theta^{\mu_n}} \mu_n (s_n^j ) \nabla_{a_n^j} Q_n (s_n^j,a_n^j,s_{ag}^j)]$;
                \item Perform gradient ascent on $\nabla_{\theta^{\mu_n}}  J({\theta_n^{\mu_n}})$ to update $\theta_n^{\mu_n}$;
                \item Update Target Networks.
            \end{itemize}
            \item After collecting $T_{horizon}$:
            \begin{itemize}[leftmargin=1em, itemsep=0pt, topsep=0pt, parsep=0pt, partopsep=0pt]
                \item Compute advantages $A_t$ and targets for value function;
                \item PPO Clipped Objective: $L^{CLIP}(\theta) = \hat{\mathbb{E}}_t \left[ \min \left( r_{ag,t}(\theta) A_t, \, clip\big(r_{ag,t}(\theta), 1 - \epsilon, 1 + \epsilon\big) A_t \right) \right]$
                \item Update Target Networks $\pi_\theta, V_\phi$ by gradient steps.
            \end{itemize}
        \end{enumerate}
    \end{enumerate}
\end{itemize}

\STATE Termination and Finalization 
\begin{itemize}[leftmargin=1em, itemsep=0pt, topsep=0pt, parsep=0pt, partopsep=0pt]
    \item $e \leftarrow e+1$ and go to 2, until  $e=e^{final}$ 
\end{itemize}

\end{algorithmic}
\end{algorithm}

\vspace{-5pt}
\section{Case Study}
The effectiveness of the proposed SEAM-LESS for coordinated wholesale and P2P retail market participation of DERs is evaluated against two rule-based (RB) and one conventional MARL market strategies:

\begin{itemize}[leftmargin=1em, itemsep=0pt, topsep=0pt, parsep=0pt, partopsep=0pt]
    \item \textit{RB Agg:} All retail energy trades are handled by the aggregator, with no P2P transactions among prosumers. The aggregator is assumed to always clear its trades in the wholesale market and sells to/buys from the prosumers at the maximum markup $\rho^{max}$.
    \item \textit{RB P2P:} All retail P2P participants settle at a uniform clearing price, providing consistent trading conditions for buyers and sellers. Any surplus or deficit after P2P trades is handled by the aggregator, who always clears in the wholesale market and sells to/buys from the prosumers at the maximum markup $\rho^{max}$.
    \item \textit{MARL-Conv:} Instead of LSD-MADDPG which maintains information privacy during centralized training, a conventional MADDPG is adopted. In this setting, all MADDPG agents share observations and actions with their centralized critic, while PPO shares only its actions.
\end{itemize}

The simulation parameters for MADDPG and LSD-MADDPG, as well as PPO are presented in Table \ref{tab_1}. Each episode was simulated over a 24-hour period, from (00:00) to (23:59), corresponding to the hourly day-ahead market.

\begin{table}[pos=b]
\centering
\footnotesize
\vspace{-15pt}
\caption{RL Training Hyperparameters}
\label{tab_1}
\renewcommand{\arraystretch}{0.95}
\begin{tabular}{l | l}
\hline
\textbf{MADDPG / LSD-MADDPG} & \textbf{PPO} \\
\hline
Total Timesteps: 60{,}000 & Total Timesteps: 60{,}000 \\
Episode Length: 24 & Episode Length: 24 \\
Learning Rate Actor: $10^{-4}$ & Learning Rate Actor: $10^{-4}$ \\
Learning Rate Critic: $10^{-3}$ & Learning Rate Critic: $10^{-3}$ \\
Noise-rate: 0.1 & GAE factor: 0.1 \\
Gamma: 0.95 & Clip rate: 0.95 \\
Tau: 0.01 & K epochs: 0.01 \\
Buffer size: $5 \times 10^5$ & L2 regularization: $10^{-3}$ \\
Batch size: 256 & Batch size: 256 \\
\hline
\end{tabular}
\vspace{-10pt}
\end{table}

Table \ref{tab_2} provides the permissible bidding specifications for the P2P and wholesale markets, including prosumers and aggregator price ranges and maximum markup. These values can be adapted to specific agreements or market rules, thereby defining the feasible cost bounds for participants in different markets.

\begin{table}[pos=h]
\centering
\footnotesize
\vspace{-10pt}
\caption{Bidding Specifications}
\label{tab_2}
\begin{tabular}{C c}
\hline
\textbf{Attributes} & \textbf{Value} \\
\hline
Prosumer Min/Max Bidding Price & $0$ -- $200\ \$/\mathrm{MWh}$ \\
Aggregator Min/Max Price Markup & $\rho^{\max} = 50\%$ \\
Aggregator Min/Max Price & $0$ -- $100\ \$/\mathrm{MWh}$ \\
\hline
\end{tabular}
\end{table}

\subsection{Effectiveness of the Proposed SEAM-LESS}
\begin{figure}
\centering
\includegraphics[width=0.48\textwidth]{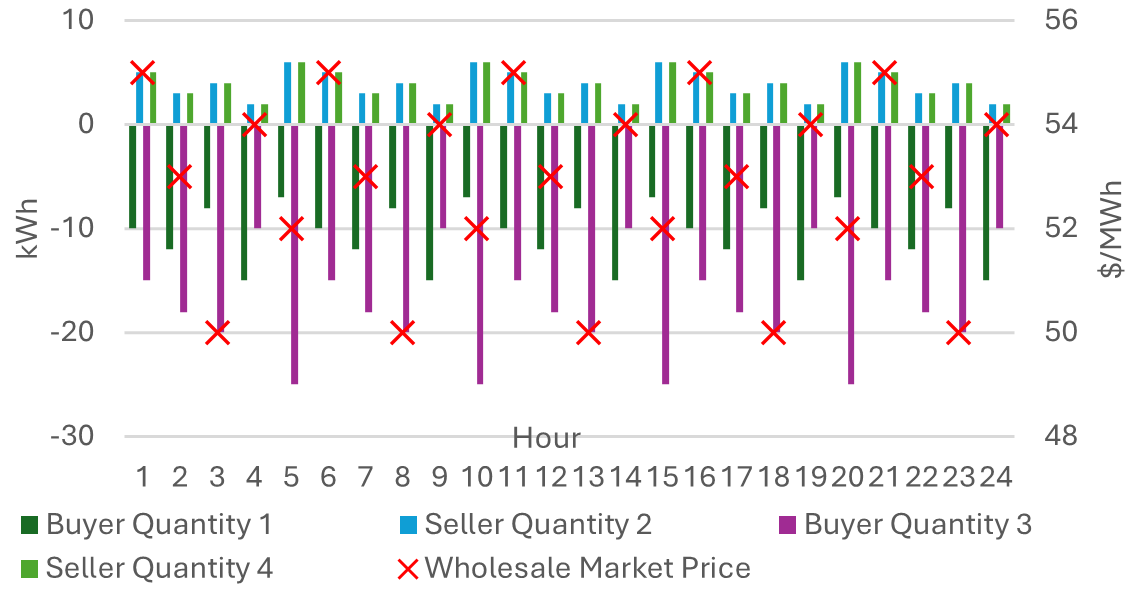}
\caption{Time-series of P2P participants energy overlaid with hourly wholesale price forecasts in Case I.}
\vspace{-15pt}
\label{fig_3}
\end{figure}

Three cases are studied to compare the four market strategies: 

Case I: Each P2P participant is designated as either a buyer or seller: prosumers 1 and 3 hold negative quantities (as buyer), while prosumers 2 and 4 always maintain positive quantities (as seller). Figure \ref{fig_3} illustrates the evaluation data, where all prosumers’ hourly quantities and wholesale prices are randomly varied by $\pm$5 units during training.

Case II: In contrast to Case I, where sellers hold market power due to scarce and valuable supply, Case II tests the inverse scenario by negating all participants’ quantities, so prosumers 1 and 3 become sellers and prosumers 2 and 4 become buyers. This setup evaluates how buyers wield market power in the P2P market with reduced demand.

Case III: This case uses PJM wholesale market data \cite{Qiu_Scalable_2021} from a 2023 year-long dataset, as well as practical PV generation and energy consumption profiles of buildings from \cite{Wilk_Multi_Agent_2024} to simulate the prosumers. Prosumers 1 and 2, with integrated solar PV systems, may switch their P2P market role from buyer to seller when their onsite PV generation exceeds energy demand during a time interval, while the remaining two prosumers remain as buyers throughout the day, demonstrating dynamic P2P interactions. Episodes are randomly sampled from yearly data to capture seasonal variations in market prices, loads, and supply conditions.

Table \ref{tab_3} compares the net monetary gains or losses for each prosumer participating in the P2P market and the aggregator in Case I. The row labeled “Sum P2P” aggregates all P2P transactions into a single net value.

\begin{table}[pos=b]
\centering
\footnotesize
\setlength{\tabcolsep}{3pt}
\renewcommand{\arraystretch}{0.9}
\vspace{-12pt}
\caption{Case I Monetary Costs and Revenue (in \$)}
\label{tab_3}
\begin{tabular}{l|c|c|c|c}
\hline
 & \textbf{RB Agg} & \textbf{RB P2P} & \textbf{MARL-Conv} & \textbf{SEAM-LESS} \\
\hline
Aggregator & 2.19  & 1.22  & 1.25  & 1.27  \\
Buyer 1    & -1.99 & -1.81 & -1.98 & -1.97 \\
Seller 2   & 0.24  & 0.50  & 0.74  & 0.69  \\
Buyer 3    & -3.21 & -2.92 & -3.29 & -3.22 \\
Seller 4   & 0.24  & 0.50  & 0.75  & 0.70  \\
\hline
Sum P2P    & -4.72 & -3.74 & -3.78 & -3.79 \\
\hline
\end{tabular}
\vspace{-10pt}
\end{table}

RB Agg is used as a benchmark to evaluate market manipulation, ensuring prices remain within acceptable limits, while RB P2P serves as a reference for a midpoint single-price market. MARL-Conv and SEAM-LESS results are comparable, both favoring sellers with market power by securing higher revenues compared to RB markets. Conversely, buyers benefit most from the RB P2P market, where a uniform P2P trading price eliminates market power disparities, resulting in the lowest buyer costs and proportional quantity splits among peers. RB Agg maximizes aggregator profit by funneling all energy through the aggregator at its bid/offer spread, elevating buyer costs and reducing seller revenues the most. Compared to RB P2P’s uniform pricing, RL-based market solutions introduce competition by enabling distinct price bids.

\begin{figure}[pos=h]
\centering
\includegraphics[width=0.48\textwidth]{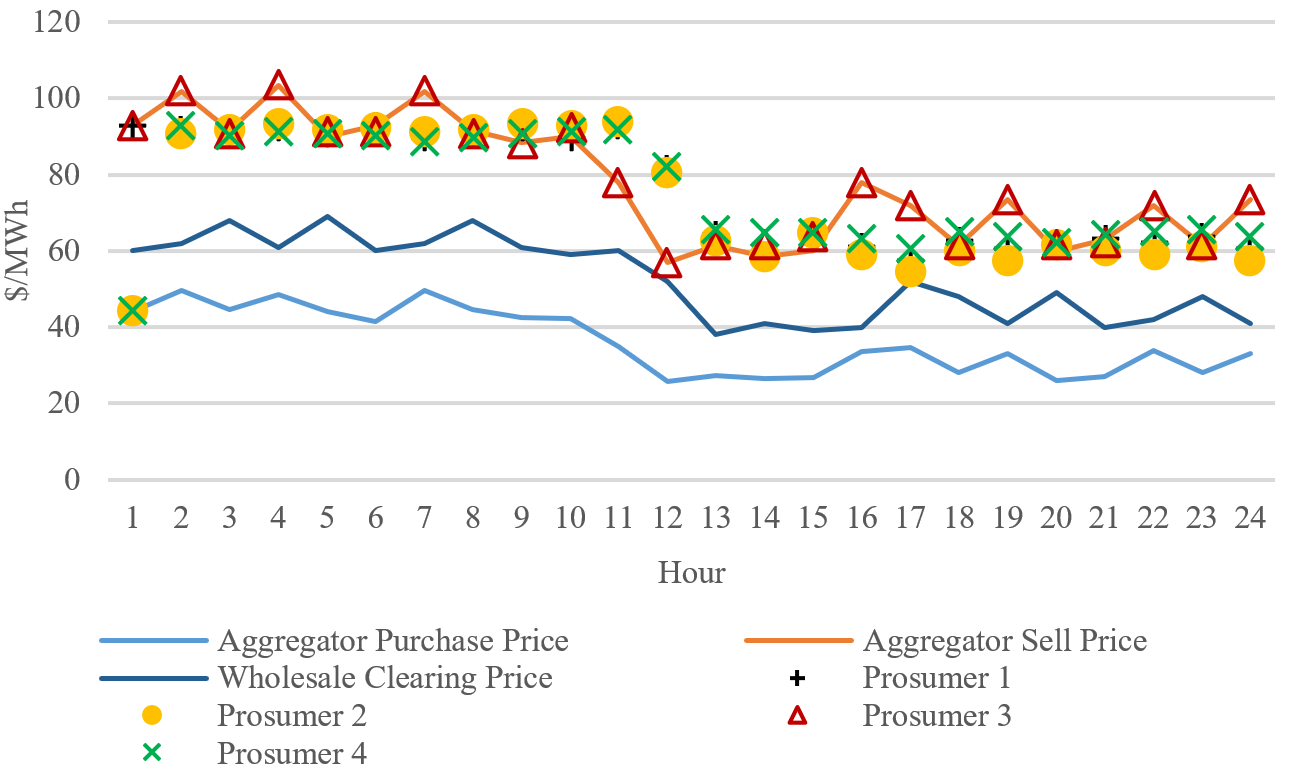}
\caption{SEAM-LESS settling price evolution of Case I.}
\label{fig_4}
\vspace{-15pt}
\end{figure}

SEAM-LESS notably enhances seller revenues while maintaining buyer costs similar to those under RB Agg, demonstrating that prosumer agents effectively exploit market power without excessive manipulation, aided by the aggregator fallback price mechanism. This is evident by the aggregator’s increased profit in SEAM-LESS over RB P2P, as mismatches in competitive pricing cause some trades to revert to the aggregator for settlement. Despite limited information sharing among prosumers, as compared to MARL-Conv, SEAM-LESS successfully learns optimal bidding strategies for prosumers. Figure. \ref{fig_4} illustrates that P2P settling prices often surpass the wholesale clearing price, boosting seller revenues while buyer costs align more closely with the aggregator’s sell price. 

As shown in Table \ref{tab_4}, SEAM-LESS yields the best Sum P2P reward (5.91), surpassing all other methods. Notably, Buyer 1 shifts from a negative reward of -0.22 to a positive reward of 0.18, and Buyer 3 increases from -0.87 to -0.47, while Seller 2 climbs above 3.14 compared to 2.60 in MARL-Conv, reflecting increased profit and lower penalties. These penalties specifically target aggregator fallback when uncompetitive prices arise, thereby guiding P2P trades learning toward stable pricing strategies.

\vspace{-5pt}
\begin{table}[pos=h]
\centering
\footnotesize
\setlength{\tabcolsep}{3pt}
\renewcommand{\arraystretch}{0.9}
\caption{Case I Rewards Comparison}
\label{tab_4}
\begin{tabular}{l | c | c | c | c}
\hline
 & \textbf{RB Agg} & \textbf{RB P2P} & \textbf{MARL-Conv} & \textbf{SEAM-LESS} \\
\hline
Aggregator & 21.99  & 12.25 & 12.57 & 12.72 \\
Buyer 1    & -13.11 & -0.81 & -0.22 & 0.18  \\
Seller 2   & -16.37 & 2.54  & 2.60  & 3.14  \\
Buyer 3    & -14.61 & 1.23  & -0.87 & -0.47 \\
Seller 4   & -16.37 & 2.54  & 3.00  & 3.05  \\
\hline
Sum P2P    & -60.45 & 5.50  & 4.50  & 5.91  \\
\hline
\end{tabular}
\end{table}

\begin{figure}[pos=h]
\centering
\includegraphics[width=0.48\textwidth]{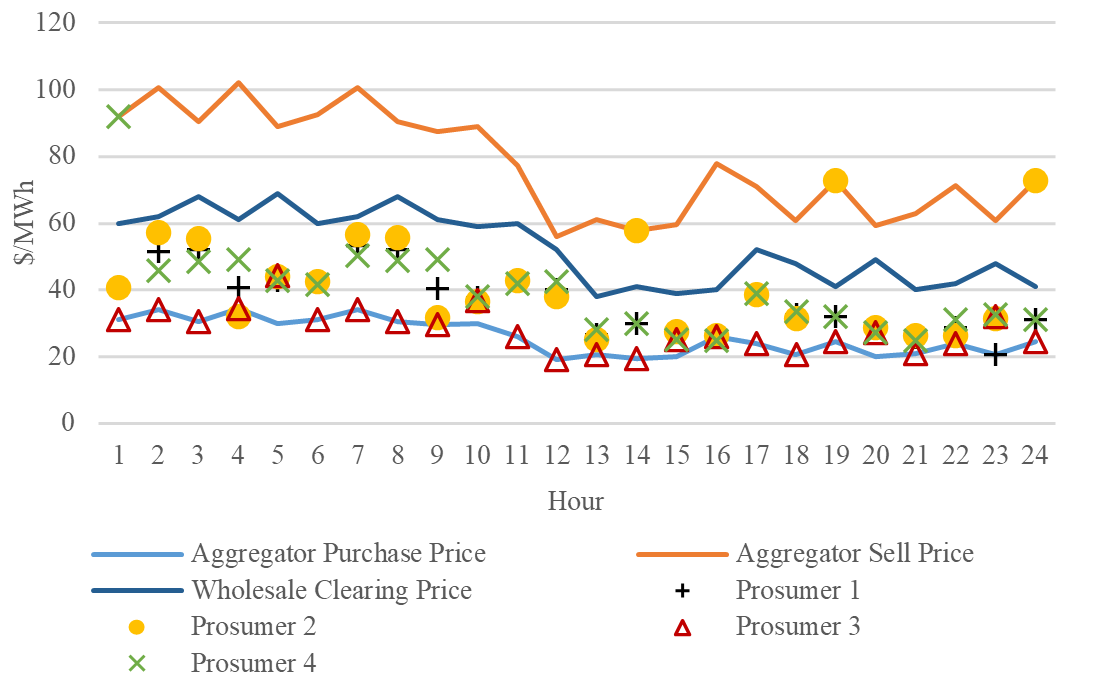}
\caption{SEAM-LESS settling price evolution of Case II.}
\vspace{-10pt}
\label{fig_5}
\end{figure}

Figures. \ref{fig_4} (for Case I) and \ref{fig_5} (for Case II) illustrate the settling price evolution of the aggregator and P2P market participants, collectively demonstrating how market power emerges when one side’s quantity is limited yet bounded by the aggregator’s fallback. In Case I, two buyers’ total demand exceeds two sellers’ total supply, forcing buyers to compete for scarce retail energy supply, driving up P2P bidding prices (and in turn retail P2P clearing prices) and weakening their bargaining position. By inverting participant quantities in Case II, the retail P2P market shifts to more supply than demand. Sellers must undercut each other to secure retail trades, lowering their offer prices and thus retail P2P clearing prices, favoring buyers with cheaper deals. Despite these shifts, neither side can push prices unboundedly high or low, as the aggregator’s purchase/sell prices ultimately constrain the P2P market power. 

\begin{figure}[pos=h]
\centering
\includegraphics[width=0.48\textwidth]{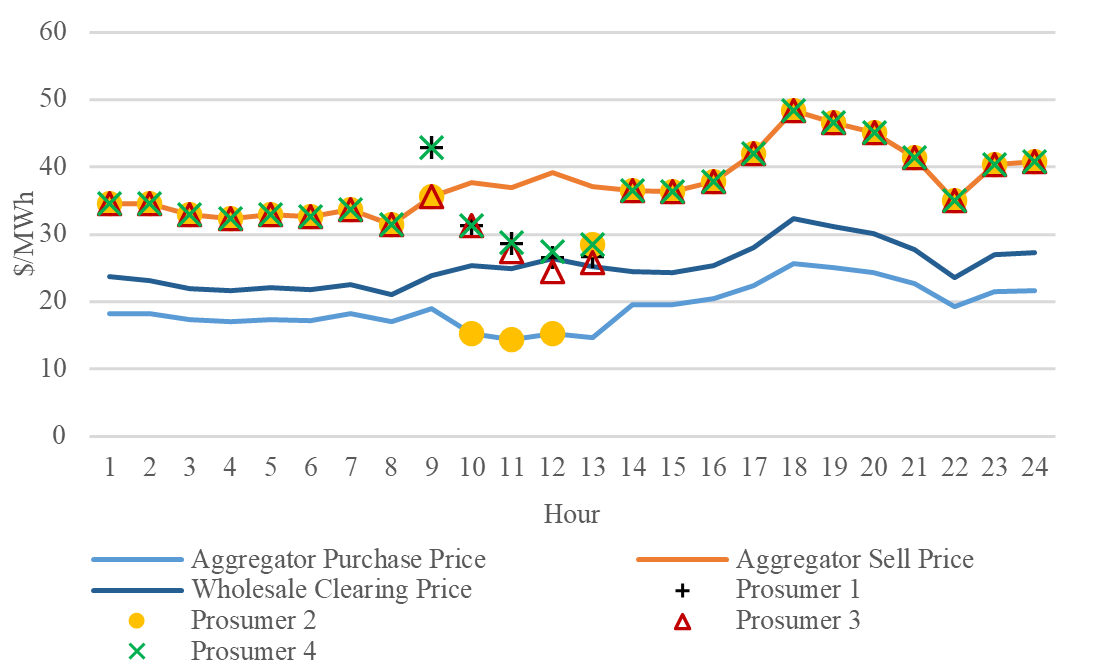}
\caption{Price dynamics with solar PV integration, highlighting elevated P2P clearing prices at peak hours.}
\vspace{-12pt}
\label{fig_6}
\end{figure}

In Case III, with the integration of solar PVs, prosumers 1 and 2’s net energy remains demand‐heavy excepting for the peak solar hours (i.e., hours 9-13), when a limited surplus starts to push up the P2P prices in hour 9, as shown in Figure. \ref{fig_6}. The onsite energy availability during these peak solar periods temporarily shifts market power and elevates P2P clearing prices in favor of sellers, but aggregator fallback quickly shapes the overall cost distribution and ultimately decreases total P2P market cost relative to the RB Aggregator approach. 

Table \ref{tab_5} presents all prosumers' monetary costs and the aggregator's revenue in Case III with MARL‐Conv yielding the highest P2P outcome, albeit with a lower aggregator revenue, while SEAM‐LESS better balances aggregator profit with competitive P2P pricing, comparable to RB P2P. Table \ref{tab_6} shows the rewards of all participants, with negative rewards reflecting missed opportunities, i.e., surplus not sold during the short peak solar window, leading to unrealized P2P transactions at favorable times. When focusing on participants’ rewards as defined in Section III and not their actual profits or costs (i.e., relative to aggregator prices $p_a^b,p_a^s$), SEAM-LESS demonstrates higher overall peer benefits. For instance, Prosumer 2, which shows a negative reward under MARL-Conv, trades more surplus in SEAM-LESS, lowering its penalty from missed transactions. Furthermore, SEAM-LESS narrows the shortfall, indicating fewer aggregator fallbacks. RB P2P and SEAM-LESS both better balance aggregator's profit with improved P2P outcomes, whereas MARL-Conv sees a larger aggregator-P2P discrepancy, as its aggregator prices yield less aggregator profit but do not necessarily maximize P2P surplus. Results confirm that solar peaks offer brief seller leverage, allowing SEAM-LESS to retain robust performance by efficiently matching generation to demand while preserving data privacy.

\begin{table}[pos=h]
\centering
\footnotesize
\setlength{\tabcolsep}{3pt}
\renewcommand{\arraystretch}{0.9}
\vspace{-10pt}
\caption{Case III Monetary Costs and Revenues (in \$)}
\label{tab_5}
\begin{tabular}{l | c | c | c | c}
\hline
 & \textbf{RB Agg} & \textbf{RB P2P} & \textbf{MARL-Conv} & \textbf{SEAM-LESS} \\
\hline
Aggregator  & 0.14  & 0.12  & 0.06  & 0.12  \\
Prosumer 1  & -0.09 & -0.08 & -0.06 & -0.07 \\
Prosumer 2  & -0.10 & -0.10 & -0.08 & -0.10 \\
Prosumer 3  & -0.02 & -0.02 & -0.02 & -0.02 \\
Prosumer 4  & -0.14 & -0.13 & -0.11 & -0.13 \\
\hline
Sum P2P     & -0.35 & -0.33 & -0.27 & -0.32 \\
\hline
\end{tabular}
\end{table}

\begin{table}[pos=h]
\centering
\footnotesize
\setlength{\tabcolsep}{3pt}
\renewcommand{\arraystretch}{0.9}
\caption{Case III Reward Comparison}
\label{tab_6}
\begin{tabular}{l | c | c | c | c}
\hline
 & \textbf{RB Agg} & \textbf{RB P2P} & \textbf{MARL-Conv} & \textbf{SEAM-LESS} \\
\hline
Aggregator  & 21.99  & 18.72 & 9.43  & 18.09 \\
Prosumer 1  & -95.45 & 3.16  & 0.49  & 5.82  \\
Prosumer 2  & -46.86 & 0.11  & -0.78 & -0.54 \\
Prosumer 3  & -3.56  & 1.00  & 0.81  & 0.92  \\
Prosumer 4  & -16.25 & 5.02  & 4.42  & 2.86  \\
\hline
Sum P2P     & -162.11 & 9.29 & 4.94  & 9.07  \\
\hline
\end{tabular}
\end{table}

Summarizing the case studies, Case I showcases a scenario of fixed retail buyers and sellers, illustrating how SEAM-LESS promotes P2P trades while balancing benefits among different parties; Case II demonstrates the scenario of shifted market power from sellers to buyers as compared to Case I, and shows that SEAM-LESS prevents extreme prices while enabling moderate leverage of such market power; Case III presents the market dynamics integrating solar PV systems, enabling prosumers to autonomously switch their roles as a seller or a buyer in the P2P market and engage in effective trading within the SEAM-LESS framework.

\subsection{Scalability of SEAM-LESS}
To further evaluate the scalability of the proposed SEAM-LESS framework, a case study with 10 prosumers was conducted, with quantities varied by ±5 units during training and evaluated on a test dataset \cite{pjm_lmps_dataminer}. The results are summarized in Table \ref{tab_7}, showing that SEAM-LESS effectively balances aggregator and prosumer benefits (compared with rule-based markets), performs comparably to MARL-Conv while preserving individual prosumers’ privacy and achieving higher P2P rewards. Compared to RB P2P, SEAM-LESS encourages price competition, yields lower total P2P rewards due to restricted data sharing and varied pricing strategies, and elevated aggregator reward. By preserving privacy, SEAM-LESS promotes solution scalability and adaptability, avoiding MARL-Conv’s centralized data issues. While this may sacrifice some collective efficiency, it is an ideal feature for deploying privacy-sensitive, resource-conservative, large-scale retail markets.

\begin{table}[pos=t]
\centering
\footnotesize
\setlength{\tabcolsep}{3pt}
\renewcommand{\arraystretch}{0.9}
\caption{Reward Comparison of 10 Prosumers Case}
\label{tab_7}
\begin{tabular}{l | c | c | c | c}
\hline
 & \textbf{RB Agg} & \textbf{RB P2P} & \textbf{MARL-Conv} & \textbf{SEAM-LESS} \\
\hline
Aggregator  & 78.92  & 9.75  & 12.91 & 11.41 \\
Prosumer 1  & -3.43  & 0.82  & 0.27  & 0.90  \\
Prosumer 2  & -5.72  & 1.43  & 0.77  & 0.69  \\
Prosumer 3  & -3.77  & 0.91  & 0.18  & 0.60  \\
Prosumer 4  & -3.40  & 0.74  & 0.39  & -0.09 \\
Prosumer 5  & -3.79  & 0.86  & -0.02 & 0.56  \\
Prosumer 6  & -10.31 & 1.43  & 0.91  & 0.72  \\
Prosumer 7  & -7.36  & 0.98  & 0.72  & 0.41  \\
Prosumer 8  & -5.61  & 0.69  & 0.65  & 0.49  \\
Prosumer 9  & -6.59  & 0.90  & 0.56  & 0.62  \\
Prosumer 10 & -8.44  & 1.14  & 0.86  & 0.63  \\
\hline
Sum P2P     & -58.4199 & 9.9025 & 5.2911 & 5.5179 \\
\hline
\end{tabular}
\vspace{-10pt}
\end{table}

\vspace{-10pt}
\section{Conclusion}
We introduced SEAM-LESS, a scalable and privacy-preserving hierarchical MARL framework for DERs' coordinated participation in wholesale and retails markets. SEAM-LESS integrates a PPO-based aggregator’s wholesale bidding and LSD-MADDPG based strategies for prosumers' retail P2P market clearing. SEAM-LESS ensures that individual prosumers adaptively align their retail market behaviors with their collective actions to advance wholesale market clearing goals through aggregator coordination. Unlike centralized training in conventional MADDPG, SEAM-LESS balances cooperative and competitive elements through localized decision-making, mitigating data bottlenecks, communication lags, and single points of failure. The case studies demonstrated SEAM-LESS’s robust performance across diverse market scenarios, highlighting its ability to coordinate DERs, aggregator fallback, and P2P trading in a stable and economically efficient manner.

Moving forward, SEAM-LESS could be extended to incorporate an even richer portfolio of DERs and to handle more complex grid constraints into the market model design, further fostering stable, economically efficient, and environmentally sustainable energy trading ecosystems in more sophisticated system contexts.

\vspace{-10pt}


\bibliographystyle{cas-model2-numbers}

\bibliography{cas-refs}

@article{BLAZQUEZ20181,
  title   = {The renewable energy policy Paradox},
  journal = {Renewable and Sustainable Energy Reviews},
  volume  = {82},
  pages   = {1-5},
  year    = {2018},
  issn    = {1364-0321},
  doi     = {https://doi.org/10.1016/j.rser.2017.09.002},
  author  = {Jorge Blazquez and Rolando Fuentes-Bracamontes and Carlo Andrea Bollino and Nora Nezamuddin}
}

@article{Wang2016,
  author  = {Wang, Hongbo},
  title   = {Do Mandatory U.S. State Renewable Portfolio Standards Increase Electricity Prices?},
  journal = {Growth and Change},
  volume  = {47},
  number  = {2},
  pages   = {157-174},
  doi     = {https://doi.org/10.1111/grow.12118},
  year    = {2016}
}

@article{greenstone2020renewable,
  title   = {Do renewable portfolio standards deliver cost-effective carbon abatement?},
  author  = {Greenstone, Michael and Nath, Ishan},
  journal = {University of Chicago, Becker Friedman Institute for Economics Working Paper. },
  number  = {2019-62},
  year    = {2020},
  url     = {https://ssrn.com/abstract=3374942}
}

@misc{statista_us_electricity_prices,
  author       = {{Statista}},
  title        = {Historical electricity prices in the United States from 1990 to 2023},
  year         = {2023},
  howpublished = {\url{https://www.statista.com}},
  note         = {Accessed: Jan. 2025}
}

@misc{rhodes_us_grid_2018,
  author      = {J. D. Rhodes},
  title       = {The old, dirty, creaky U.S. electric grid would cost \$5 trillion to replace. Where should infrastructure spending go?},
  year        = {2018},
  month       = dec,
  institution = {Energy Institute, The University of Texas at Austin},
  url         = {https://energy.utexas.edu/news/old-dirty-creaky-us-electric-grid-would-cost-5-trillion-replace-where-should-infrastructure},
  note        = {Accessed: Jan. 2025}
}

@article{strielkowski2021renewable,
  author         = {Strielkowski, Wadim and Civín, Lubomír and Tarkhanova, Elena and Tvaronavičienė, Manuela and Petrenko, Yelena},
  title          = {Renewable Energy in the Sustainable Development of Electrical Power Sector: A Review},
  journal        = {Energies},
  volume         = {14},
  year           = {2021},
  number         = {24},
  article-number = {8240},
  issn           = {1996-1073},
  doi            = {10.3390/en14248240}
}

@article{hahnel2020becoming,
  title   = {Becoming prosumer: Revealing trading preferences and decision-making strategies in peer-to-peer energy communities},
  journal = {Energy Policy},
  volume  = {137},
  pages   = {111098},
  year    = {2020},
  issn    = {0301-4215},
  doi     = {https://doi.org/10.1016/j.enpol.2019.111098},
  author  = {Ulf J.J. Hahnel and Mario Herberz and Alejandro Pena-Bello and David Parra and Tobias Brosch}
}

@book{pinto_local_electricity_markets_2021,
  author    = {T. Pinto and Z. Vale and S. Widergren},
  title     = {Local Electricity Markets},
  publisher = {Elsevier},
  address   = {Amsterdam, The Netherlands},
  year      = {2021},
  isbn      = {978-0-12-820074-2},
  url       = {https://www.sciencedirect.com/book/edited-volume/9780128200742/local-electricity-markets}
}

@article{Ye_Multi_Agent_2023,
  author  = {Ye, Yujian and Papadaskalopoulos, Dimitrios and Yuan, Quan and Tang, Yi and Strbac, Goran},
  journal = {IEEE Transactions on Smart Grid},
  title   = {Multi-Agent Deep Reinforcement Learning for Coordinated Energy Trading and Flexibility Services Provision in Local Electricity Markets},
  year    = {2023},
  volume  = {14},
  number  = {2},
  pages   = {1541-1554},
  doi     = {10.1109/TSG.2022.3149266}
}

@article{Chen_Indirect_2019,
  author  = {Chen, Tao and Su, Wencong},
  journal = {IEEE Transactions on Smart Grid},
  title   = {Indirect Customer-to-Customer Energy Trading With Reinforcement Learning},
  year    = {2019},
  volume  = {10},
  number  = {4},
  pages   = {4338-4348},
  doi     = {10.1109/TSG.2018.2857449}
}

@article{Strbac_Decarbonization_2021,
  author    = {Strbac, Goran and Papadaskalopoulos, Dimitrios and Chrysanthopoulos, Nikolaos and Estanqueiro, Ana and Algarvio, Hugo and Lopes, Fernando and de Vries, Laurens and Morales-Espana, German and Sijm, Jos and Hernandez-Serna, Ricardo and Kiviluoma, Juha and Helisto, Niina},
  address   = {[New York] :},
  issn      = {1540-7977},
  journal   = {IEEE power \& energy magazine},
  lccn      = {2003263905},
  number    = {1},
  publisher = {Institute of Electrical and Electronics Engineers :},
  title     = {Decarbonization of Electricity Systems in Europe: Market Design Challenges},
  volume    = {19},
  year      = {2021-1},
  doi       = {10.1109/MPE.2020.3033397}
}

@article{Qiu_Scalable_2021,
  title   = {Scalable coordinated management of peer-to-peer energy trading: A multi-cluster deep reinforcement learning approach},
  journal = {Applied Energy},
  volume  = {292},
  pages   = {116940},
  year    = {2021},
  issn    = {0306-2619},
  doi     = {https://doi.org/10.1016/j.apenergy.2021.116940},
  author  = {Dawei Qiu and Yujian Ye and Dimitrios Papadaskalopoulos and Goran Strbac}
}

@inproceedings{Ghasemi_Mult_Agent_2020,
  author    = {Ghasemi, Arman and Shojaeighadikolaei, Amin and Jones, Kailani and Hashemi, Morteza and Bardas, Alexandru G. and Ahmadi, Reza},
  booktitle = {2020 IEEE International Conference on Communications, Control, and Computing Technologies for Smart Grids (SmartGridComm)},
  title     = {A Multi-Agent Deep Reinforcement Learning Approach for a Distributed Energy Marketplace in Smart Grids},
  year      = {2020},
  volume    = {},
  number    = {},
  pages     = {1-6},
  doi       = {10.1109/SmartGridComm47815.2020.9302981}
}

@article{Papadaskalopoulos_Nonlinear_2016,
  author   = {Papadaskalopoulos, Dimitrios and Strbac, Goran},
  journal  = {IEEE Transactions on Smart Grid},
  title    = {Nonlinear and Randomized Pricing for Distributed Management of Flexible Loads},
  year     = {2016},
  volume   = {7},
  number   = {2},
  pages    = {1137-1146},
  doi      = {10.1109/TSG.2015.2437795}
}

@article{Yang_Deep_2022,
  author  = {Yang, Guang and Du, Songhuai and Duan, Qingling and Su, Juan},
  title   = {Deep Reinforcement Learning-Based Trading Strategy for Load Aggregators on Price-Responsive Demand},
  journal = {Computational Intelligence and Neuroscience},
  volume  = {2022},
  number  = {1},
  pages   = {6884956},
  doi     = {https://doi.org/10.1155/2022/6884956},
  year    = {2022}
}

@article{Le_Ray_Evaluating_2018,
  author   = {Le Ray, Guillaume and Larsen, Emil Mahler and Pinson, Pierre},
  journal  = {IEEE Transactions on Smart Grid},
  title    = {Evaluating Price-Based Demand Response in Practice—With Application to the EcoGrid EU Experiment},
  year     = {2018},
  volume   = {9},
  number   = {3},
  pages    = {2304-2313},
  doi      = {10.1109/TSG.2016.2610518}
}

@article{Khojasteh_A_Novel_2023,
  title   = {A novel adaptive robust model for scheduling distributed energy resources in local electricity and flexibility markets},
  journal = {Applied Energy},
  volume  = {342},
  pages   = {121144},
  year    = {2023},
  issn    = {0306-2619},
  doi     = {https://doi.org/10.1016/j.apenergy.2023.121144},
  author  = {Meysam Khojasteh and Pedro Faria and Fernando Lezama and Zita Vale}
}

@article{Wang_Incentivizing_2018,
  author  = {Wang, Hao and Huang, Jianwei},
  journal = {IEEE Transactions on Smart Grid},
  title   = {Incentivizing Energy Trading for Interconnected Microgrids},
  year    = {2018},
  volume  = {9},
  number  = {4},
  pages   = {2647-2657},
  doi     = {10.1109/TSG.2016.2614988}
}

@inproceedings{Agwan_Pricing_2021,
  author    = {Agwan, Utkarsha and Spangher, Lucas and Arnold, William and Srivastava, Tarang and Poolla, Kameshwar and Spanos, Costas J.},
  title     = {Pricing in Prosumer Aggregations using Reinforcement Learning},
  year      = {2021},
  isbn      = {9781450383332},
  publisher = {Association for Computing Machinery},
  address   = {New York, NY, USA},
  doi       = {10.1145/3447555.3464853},
  pages     = {220–224},
  numpages  = {5},
  location  = {Virtual Event, Italy},
  series    = {e-Energy '21}
}

@inproceedings{Jia_Retail_2013,
  author    = {Jia, Liyan and Zhao, Qing and Tong, Lang},
  booktitle = {2013 51st Annual Allerton Conference on Communication, Control, and Computing (Allerton)},
  title     = {Retail pricing for stochastic demand with unknown parameters: An online machine learning approach},
  year      = {2013},
  volume    = {},
  number    = {},
  pages     = {1353-1358},
  doi       = {10.1109/Allerton.2013.6736684}
}

@article{Kim_An_Online_2017,
  author  = {Kim, Seung-Jun and Giannakis, Geogios B.},
  journal = {IEEE Transactions on Smart Grid},
  title   = {An Online Convex Optimization Approach to Real-Time Energy Pricing for Demand Response},
  year    = {2017},
  volume  = {8},
  number  = {6},
  pages   = {2784-2793},
  doi     = {10.1109/TSG.2016.2539948}
}

@article{Liu_Energy_Sharing_2017,
  author   = {Liu, Nian and Yu, Xinghuo and Wang, Cheng and Li, Chaojie and Ma, Li and Lei, Jinyong},
  journal  = {IEEE Transactions on Power Systems},
  title    = {Energy-Sharing Model With Price-Based Demand Response for Microgrids of Peer-to-Peer Prosumers},
  year     = {2017},
  volume   = {32},
  number   = {5},
  pages    = {3569-3583},
  doi      = {10.1109/TPWRS.2017.2649558}
}

@article{Horrillo_Quintero_Smart_2026,
  title   = {Smart energy coordination in microgrid clusters using hybrid model predictive control and differential evolution optimization},
  journal = {Energy Conversion and Management},
  volume  = {351},
  pages   = {121039},
  year    = {2026},
  issn    = {0196-8904},
  doi     = {https://doi.org/10.1016/j.enconman.2026.121039},
  author  = {Pablo Horrillo-Quintero and Pablo García-Triviño and David Carrasco-González and Carlos Andrés García-Vázquez and Luis M. Fernández-Ramírez}
}

@article{Vinecent_Pastor_Evaluation_2019,
  author   = {Vicente-Pastor, Alejandro and Nieto-Martin, Jesus and Bunn, Derek W. and Laur, Arnaud},
  journal  = {IEEE Transactions on Power Systems},
  title    = {Evaluation of Flexibility Markets for Retailer–DSO–TSO Coordination},
  year     = {2019},
  volume   = {34},
  number   = {3},
  pages    = {2003-2012},
  doi      = {10.1109/TPWRS.2018.2880123}
}

@article{Khajeh_Local_Capacity_2021,
  author   = {Khajeh, Hosna and Firoozi, Hooman and Hesamzadeh, Mohammad Reza and Laaksonen, Hannu and Shafie-Khah, Miadreza},
  journal  = {IEEE Access},
  title    = {A Local Capacity Market Providing Local and System-Wide Flexibility Services},
  year     = {2021},
  volume   = {9},
  number   = {},
  pages    = {52336-52351},
  doi      = {10.1109/ACCESS.2021.3069949}
}

@article{Zhou_Framework_2020,
  title   = {Framework design and optimal bidding strategy for ancillary service provision from a peer-to-peer energy trading community},
  journal = {Applied Energy},
  volume  = {278},
  pages   = {115671},
  year    = {2020},
  issn    = {0306-2619},
  doi     = {https://doi.org/10.1016/j.apenergy.2020.115671},
  author  = {Yue Zhou and Jianzhong Wu and Guanyu Song and Chao Long}
}

@article{Guo_Change_Constrainted_2021,
  author   = {Guo, Zhenwei and Pinson, Pierre and Chen, Shibo and Yang, Qinmin and Yang, Zaiyue},
  journal  = {IEEE Transactions on Smart Grid},
  title    = {Chance-Constrained Peer-to-Peer Joint Energy and Reserve Market Considering Renewable Generation Uncertainty},
  year     = {2021},
  volume   = {12},
  number   = {1},
  pages    = {798-809},
  doi      = {10.1109/TSG.2020.3019603}
}

@article{Cao_Reinforcement_2020,
  author   = {Cao, Di and Hu, Weihao and Zhao, Junbo and Zhang, Guozhou and Zhang, Bin and Liu, Zhou and Chen, Zhe and Blaabjerg, Frede},
  journal  = {Journal of Modern Power Systems and Clean Energy},
  title    = {Reinforcement Learning and Its Applications in Modern Power and Energy Systems: A Review},
  year     = {2020},
  volume   = {8},
  number   = {6},
  pages    = {1029-1042},
  doi      = {10.35833/MPCE.2020.000552}
}

@book{sutton_reinforcement_learning_2018,
  author    = {Richard S. Sutton and Andrew G. Barto},
  title     = {Reinforcement Learning: An Introduction},
  publisher = {MIT Press},
  address   = {Cambridge, MA, USA},
  year      = {2018},
  edition   = {2nd},
  isbn      = {9780262039246},
  url       = {http://incompleteideas.net/book/the-book-2nd.html}
}

@article{Chen_Local_2018,
  author  = {Chen, Tao and Su, Wencong},
  journal = {IEEE Access},
  title   = {Local Energy Trading Behavior Modeling With Deep Reinforcement Learning},
  year    = {2018},
  volume  = {6},
  number  = {},
  pages   = {62806-62814},
  doi     = {10.1109/ACCESS.2018.2876652}
}

@article{Wan_Model_Free_2019,
  author   = {Wan, Zhiqiang and Li, Hepeng and He, Haibo and Prokhorov, Danil},
  journal  = {IEEE Transactions on Smart Grid},
  title    = {Model-Free Real-Time EV Charging Scheduling Based on Deep Reinforcement Learning},
  year     = {2019},
  volume   = {10},
  number   = {5},
  pages    = {5246-5257},
  doi      = {10.1109/TSG.2018.2879572}
}

@article{Hua_Optimal_2019,
  title   = {Optimal energy management strategies for energy Internet via deep reinforcement learning approach},
  journal = {Applied Energy},
  volume  = {239},
  pages   = {598-609},
  year    = {2019},
  issn    = {0306-2619},
  doi     = {https://doi.org/10.1016/j.apenergy.2019.01.145},
  author  = {Haochen Hua and Yuchao Qin and Chuantong Hao and Junwei Cao}
}

@article{Brandi_Deep_Reinforcement_2020,
  title   = {Deep reinforcement learning to optimise indoor temperature control and heating energy consumption in buildings},
  journal = {Energy and Buildings},
  volume  = {224},
  pages   = {110225},
  year    = {2020},
  issn    = {0378-7788},
  doi     = {https://doi.org/10.1016/j.enbuild.2020.110225},
  author  = {Silvio Brandi and Marco Savino Piscitelli and Marco Martellacci and Alfonso Capozzoli}
}

@article{Anvari_Moghaddam_A_Multi_Agent_2017,
  title   = {A multi-agent based energy management solution for integrated buildings and microgrid system},
  journal = {Applied Energy},
  volume  = {203},
  pages   = {41-56},
  year    = {2017},
  issn    = {0306-2619},
  doi     = {https://doi.org/10.1016/j.apenergy.2017.06.007},
  author  = {Amjad Anvari-Moghaddam and Ashkan Rahimi-Kian and Maryam S. Mirian and Josep M. Guerrero}
}

@article{Kim_Automatic_2020,
  author         = {Kim, Jin-Gyeom and Lee, Bowon},
  title          = {Automatic P2P Energy Trading Model Based on Reinforcement Learning Using Long Short-Term Delayed Reward},
  journal        = {Energies},
  volume         = {13},
  year           = {2020},
  number         = {20},
  article-number = {5359},
  issn           = {1996-1073},
  doi            = {10.3390/en13205359}
}

@misc{Hernandez_Leal_A_Survey_2019,
  title        = {A Survey of Learning in Multiagent Environments: Dealing with Non-Stationarity},
  author       = {Pablo Hernandez-Leal and Michael Kaisers and Tim Baarslag and Enrique Munoz de Cote},
  year         = {2019},
  primaryclass = {cs.MA},
  url          = {https://arxiv.org/abs/1707.09183}
}

@article{Vazquez_Canteli_Multi_Agent_2019,
  doi       = {10.1088/1742-6596/1343/1/012058},
  year      = {2019},
  month     = {nov},
  publisher = {IOP Publishing},
  volume    = {1343},
  number    = {1},
  pages     = {012058},
  author    = {Vazquez-Canteli, Jose and Detjeen, Thomas and Henze, Gregor and Kämpf, Jérôme and Nagy, Zoltan},
  title     = {Multi-agent reinforcement learning for adaptive demand response in smart cities},
  journal   = {Journal of Physics: Conference Series}
}

@article{Lu_Multi_Agent_2020,
  title   = {Multi-agent deep reinforcement learning based demand response for discrete manufacturing systems energy management},
  journal = {Applied Energy},
  volume  = {276},
  pages   = {115473},
  year    = {2020},
  issn    = {0306-2619},
  doi     = {https://doi.org/10.1016/j.apenergy.2020.115473},
  author  = {Renzhi Lu and Yi-Chang Li and Yuting Li and Junhui Jiang and Yuemin Ding}
}

@article{Wilk_Multi_Agent_2024,
  author         = {Wilk, Patrick and Wang, Ning and Li, Jie},
  title          = {Multi-Agent Reinforcement Learning for Smart Community Energy Management},
  journal        = {Energies},
  volume         = {17},
  year           = {2024},
  number         = {20},
  article-number = {5211},
  issn           = {1996-1073},
  doi            = {10.3390/en17205211}
}

@misc{pjm_lmps_dataminer,
  author       = {{PJM Interconnection}},
  title        = {{PJM Data Miner - Settlements Verified Hourly LMPs}},
  year         = {2025},
  howpublished = {\url{https://dataminer2.pjm.com/feed/rt_da_monthly_lmps}},
  note         = {Accessed: Jan. 19, 2025}
}

\end{document}